\documentclass{article}
\usepackage{microtype}
\usepackage{graphicx}
\usepackage{subfigure}
\usepackage{booktabs} 

\usepackage[ruled,vlined]{algorithm2e}

\usepackage{hyperref}

\usepackage{amsmath}
\renewcommand\vec{\mathbf}

\usepackage{easyReview} 

\usepackage[accepted]{icml2020}

\icmltitlerunning{Adaptive Braking for Mitigating Gradient Delay}

\begin{document}

\twocolumn[
\icmltitle{Adaptive Braking for Mitigating Gradient Delay}



\icmlsetsymbol{equal}{*}

\begin{icmlauthorlist}
\icmlauthor{Abhinav Venigalla}{equal,cb}
\icmlauthor{Atli Kosson}{equal,cb}
\icmlauthor{Vitaliy Chiley}{cb}
\icmlauthor{Urs K{\"o}ster}{g}

\end{icmlauthorlist}

\icmlaffiliation{cb}{Cerebras Systems, Los Altos, CA}
\icmlaffiliation{g}{Google, San Diego, CA, work done while at Cerebras Systems}

\icmlcorrespondingauthor{Abhinav Venigalla}{abhi@cerebras.net}
\icmlcorrespondingauthor{Atli Kosson}{atli@cerebras.net}

\icmlkeywords{Machine Learning, ICML, Adaptive Braking, Gradient Delay, ASGD, Gradient Staleness, Deep Learning, Distributed Training}

\vskip 0.3in
]



\printAffiliationsAndNotice{\icmlEqualContribution} 

\begin{abstract}
Neural network training is commonly accelerated by using multiple synchronized workers to compute gradient updates in parallel. Asynchronous methods remove synchronization overheads and improve hardware utilization at the cost of introducing gradient delay, which impedes optimization and can lead to lower final model performance. We introduce Adaptive Braking (AB), a modification for momentum-based optimizers that mitigates the effects of gradient delay. AB dynamically scales the gradient based on the alignment of the gradient and the velocity. This can dampen oscillations along high curvature directions of the loss surface, stabilizing and accelerating asynchronous training. We show that applying AB on top of SGD with momentum enables training ResNets on CIFAR-10 and ImageNet-1k with delays exceeding 32 update steps with minimal drop in final test accuracy.
\end{abstract}

\section{Introduction}
Computational workloads for training state-of-the-art deep learning models have grown rapidly in recent years \cite{amodei2018ai}.
This growth has outpaced the growth in compute power available on individual accelerators. To keep training times manageable, these workloads are often distributed over a cluster of devices working in parallel.
The most common form of distributed training is Distributed Synchronized SGD \cite{chen2016revisiting} which divides a mini-batch of samples between workers and accumulates the gradients from all workers before updating the model parameters.
The workers are synchronized and can not start processing the next mini-batch until the weights have been updated which lowers hardware utilization.

\citeauthor{lian2015asynchronous}~\citeyearpar{lian2015asynchronous} propose performing asynchronous weight updates to avoid the synchronization overhead.
Asynchronous weight updates can improve hardware utilization at the cost of introducing gradient delay \cite{lian2015asynchronous}.
The effects of gradient delay have been studied in several works.
\citeauthor{yang2019pipemare}~\citeyearpar{yang2019pipemare} show that delays cause unstable oscillations in the optimization trajectory lowering the maximum stable learning rate.
\citeauthor{mitliagkas2016asynchrony}~\citeyearpar{mitliagkas2016asynchrony} show that delays with a particular distribution can increase the effective momentum in the underlying optimization process.
\citeauthor{giladi2019stabilitysedge}~\citeyearpar{giladi2019stabilitysedge} and \citeauthor{kosson2020pipelined}~\citeyearpar{kosson2020pipelined} suggest that gradient delay removes the benefits of momentum and that with delays momentum should only be used if modified.
Without mitigation, gradient delay commonly results in slower optimization and worse final model performance \cite{chen2016revisiting}.
Many methods have been proposed to improve convergence with gradient delay \cite{hakimi2019dana,zhang2016staleness,zheng2017asynchronous,gaun2017delay,rigazzi2019dc3sgd,giladi2019stabilitysedge}.

\textit{Adaptive Braking} (AB) is a modification to the momentum update process that can greatly increase tolerance to delayed gradients with minimal compute overhead and no memory overhead.
AB dynamically scales the gradient magnitude based on the angle between the gradient and velocity vectors, decreasing it for positive alignment (acute angle) and increasing it for negative alignment (obtuse angle).
Intuitively AB can dampen oscillations along a single gradient component by reducing the velocity magnitude at every step.
In the case of multiple components with different, constant, curvatures, the alignment of the gradient and velocity will be more strongly correlated with the high curvature components.
This means that AB primarily dampens oscillations for the components with high curvature, stabilizing them without affecting the other components as much on average.
This resembles the effect of higher order optimization methods which account for the loss landscape curvature.

In this work we focus on applying AB to Stochastic Gradient Descent with Momentum (SGDM).
We show that training with SGDM+AB can improve asynchronous multi-worker training in multiple settings with no tuning of the single-worker hyperparameters.
In particular, SGDM+AB enables training ResNet-20 on CIFAR-10 and ResNet-50 on ImageNet-1k with large delays $D \geq 32$ with minimal accuracy degradation. 
In our experiments we compare AB with several other mitigation methods showing that AB enables greater delay tolerance than other methods.

\section{Algorithm}
Adaptive Braking (AB) is a general technique for momentum-based optimizers. It computes a gradient-velocity alignment score and uses it to scale the gradient. In this section we describe how AB is applied to SGDM.

The original SGDM update is:
\begin{align}
    \vec{v}_{t+1} &= m \vec{v}_t + \vec{g}_{t}  \label{eq:sgd_v_updt} \\
    \vec{w}_{t+1} &= \vec{w}_t - \eta \vec{v}_{t+1} \label{eq:sgd_w_updt}
\end{align}

where $\vec{w}_t$ and $\vec{v}_t$ are the model weights and velocity at time $t$, $\eta$ is the learning rate, and $m$ is the momentum coefficient. The weight gradient applied at time $t$ is $\vec{g}_{t}$ which may have been computed with a delay, $\vec{g}_{t} = G(\vec{w}_{t-D})$, where $G(\cdot)$ is the gradient function and $D$ is a random variable representing the system delay.

In \eqref{eq:sgd_v_updt} and \eqref{eq:sgd_w_updt}, each weight parameter is independent and can be processed separately. Adaptive Braking groups parameters so that it can compute a gradient-velocity alignment per group. By default we use a filter-wise grouping of parameters as described in Appendix \ref{sec:param-grouping}.

To apply Adaptive Braking, we compute the gradient scaling factor $\alpha^i_t$ based on the cosine similarity of the velocity and gradient vectors for parameter group $i$. The SGDM+AB update is:

\begin{align}
    \alpha^i_{t} &= 1 - \rho \frac{\langle \vec{g}^i_{t}, \vec{v}^i_t \rangle}{\max(\|\vec{g}^i_{t}\| \|\vec{v}^i_t\|, \epsilon)} \label{eq:ab_alpha}\\
    &\approx 1 - \rho \cos \angle \left( \vec{g}^i_{t}, \vec{v}^i_t \right) \nonumber \\    
    \vec{v}^i_{t+1}&= m \vec{v}^i_t + \alpha^i_t \vec{g}^i_{t}  \label{eq:ab_v_updt} \\
    \vec{w}^i_{t+1} &= \vec{w}^i_t - \eta \vec{v}^i_{t+1} \label{eq:ab_w_updt}
\end{align}

where $\rho$ is a scalar hyperparameter we call the \emph{braking coefficient} (Appendix~\ref{sec:alpha}), and $\epsilon$ is used for numerical stability. Different formulations of Adaptive Braking could substitute the cosine similarity with other distance functions. 

\section{Optimizing a Noisy Quadratic Model}
\label{sec:nqm}
To gain insights into how AB can help optimization, we analyze its effect on convergence for a Noisy Quadratic Model (NQM).
We adopt the setup that \citeauthor{zhang2019algorithmic}~\citeyearpar{zhang2019algorithmic} used to model the effects of batch size in neural networks.
The NQM allows us to explicitly control various aspects of the optimization such as the dimensionality, the amount of noise, the condition number, and the delay.
We measure the quality of optimization trajectories by the number of optimization steps, $T$, required to reach a target loss.
See Appendix~\ref{sec:nqm_setup} for details about our setup.

\begin{figure}[t]
    \vskip -0.05in
    \centering
    \includegraphics[width=\linewidth]{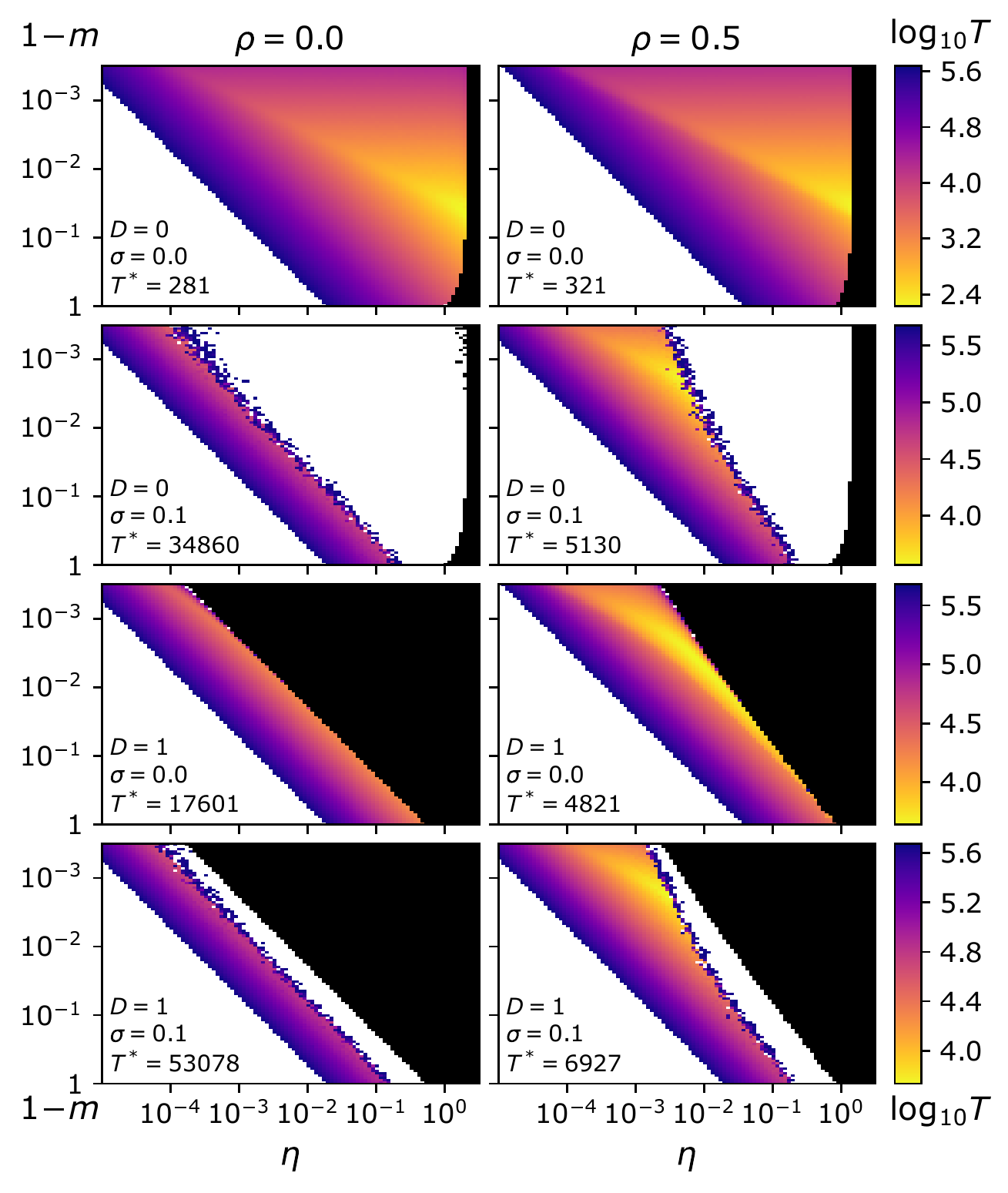}
    \vskip -0.175in
    \caption{
    This figure shows the number of optimization steps, $T$, required to reach the target loss on the NQM from Section~\ref{sec:nqm} for different hyperparameters. Each heatmap plots $T$ over different learning rates $\eta$ and momentum $m$. Black regions are unstable and white regions to not reach the target loss within the 500000 steps performed. The left column shows SGDM and the right column shows Adaptive Braking with $\rho=0.5$. The rows show different amounts of delay $D$ and noise $\sigma$. $T^*$ estimates the fastest trajectory based on the 1st percentile of $T$ over the colored region (to reduce the effects of noise and the choice of sampling grid).
    }
    \vskip -0.1in
    \label{fig:heatmap}
\end{figure}

Figure~\ref{fig:heatmap} compares $T$ for SGDM with and without AB for different learning rates, momentum values, delay and noise.
The first row shows the no-delay and no-noise case.
In this case AB does not improve the speed and slightly decreases the highest stable learning rate.
This happens because AB can magnify certain high-frequency oscillations, where $g$ and $v$ are almost always oppositely aligned, causing AB to effectively scale the learning rate by up to $1+\rho$. Appendix~\ref{sec:nqm_microstepping} explores this effect further and shows how AB can be modified to avoid this.

\begin{figure}[t!]
    \vskip -0.05in
    \centering
    \includegraphics[width=\linewidth]{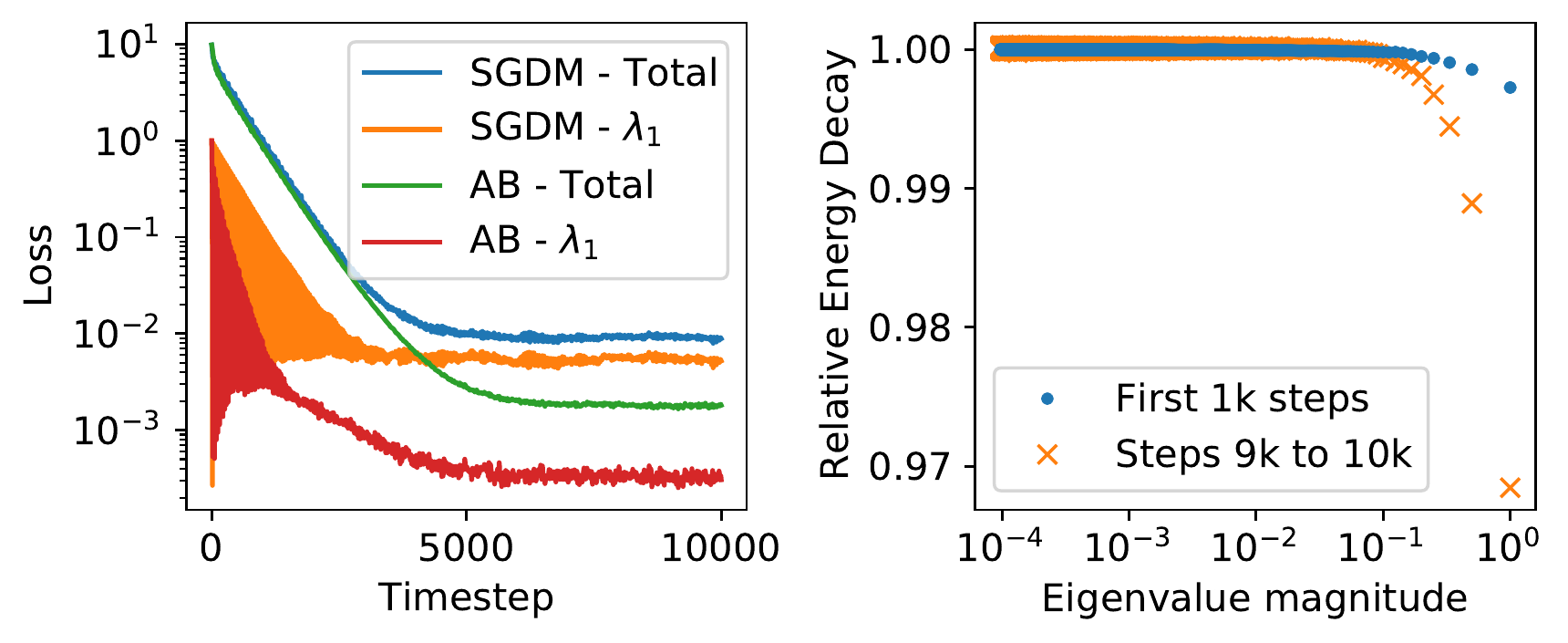}
    \caption{
    AB can lower the steady state loss when optimizing a noisy quadratic model.
    The learning rate and momentum correspond to values that reach the target loss ($10^{-2}$) with AB but not SGDM in the second row of Figure~\ref{fig:heatmap}.
    \textbf{Left:} The total loss in each case and the contribution to the loss from the largest eigenvalue.
    The steady-state loss for the largest eigenvalues is lower with AB.
    \textbf{Right:} The relative energy decay ratio for each eigenvalue showing a greater dampening of high curvature components.
    }
    \label{fig:cq_dampening_noise}
\end{figure}

The results of adding noise in the no-delay case are shown in the second row of Figure~\ref{fig:heatmap}.
In this case AB significantly speeds up the fastest trajectory and expands the region that reaches the target loss within the step budget.
In the presence of noise, a constant learning rate trajectory will converge to an expected steady-state loss that depends on the hyperparameters and level of noise. \citeauthor{zhang2019algorithmic}~\citeyearpar{zhang2019algorithmic} show that there is a trade-off with increasing the momentum and/or learning rate: it can improve the convergence rate (of the expectation) but magnifies the steady-state loss.
The dampening effect of AB can reduce the steady-state loss, expanding the region that will converge within the time limit and unlocking the faster trajectories with larger step sizes.
To measure the dampening effect of AB we can compare the energy after making an AB update ($E_{t+1}$) to what the energy would have been after making an SGDM update ($\hat{E}_{t+1})$ from each state ($\vec{w}_t$, $\vec{v}_t$) along the AB optimization trajectory. 
The energy $E_t=\mathcal{L}(\vec{w}_t) + \frac{1}{2}\eta \|\vec{v}_t\|^2$ accounts for both potential energy (the loss $\mathcal{L}(\vec{w}_t)$) and kinetic energy $\frac{1}{2}\eta \|\vec{v}\|_t^2$ of an optimization state (see Appendix~\ref{sec:nqm_energy}).
The geometric mean of $E_{t+1}/\hat{E}_{t+1}$, which we call the \emph{relative energy decay}, indicates how much faster AB dissipates energy compared to SGDM on average.
The relative energy decay can be computed for each eigenvector to measure the dampening for different components.
Figure~\ref{fig:cq_dampening_noise} shows that AB can lower the steady state-loss by dampening the large eigenvalue components.

\begin{figure}[t!]
    \vskip -0.05in
    \centering
    \includegraphics[width=\linewidth]{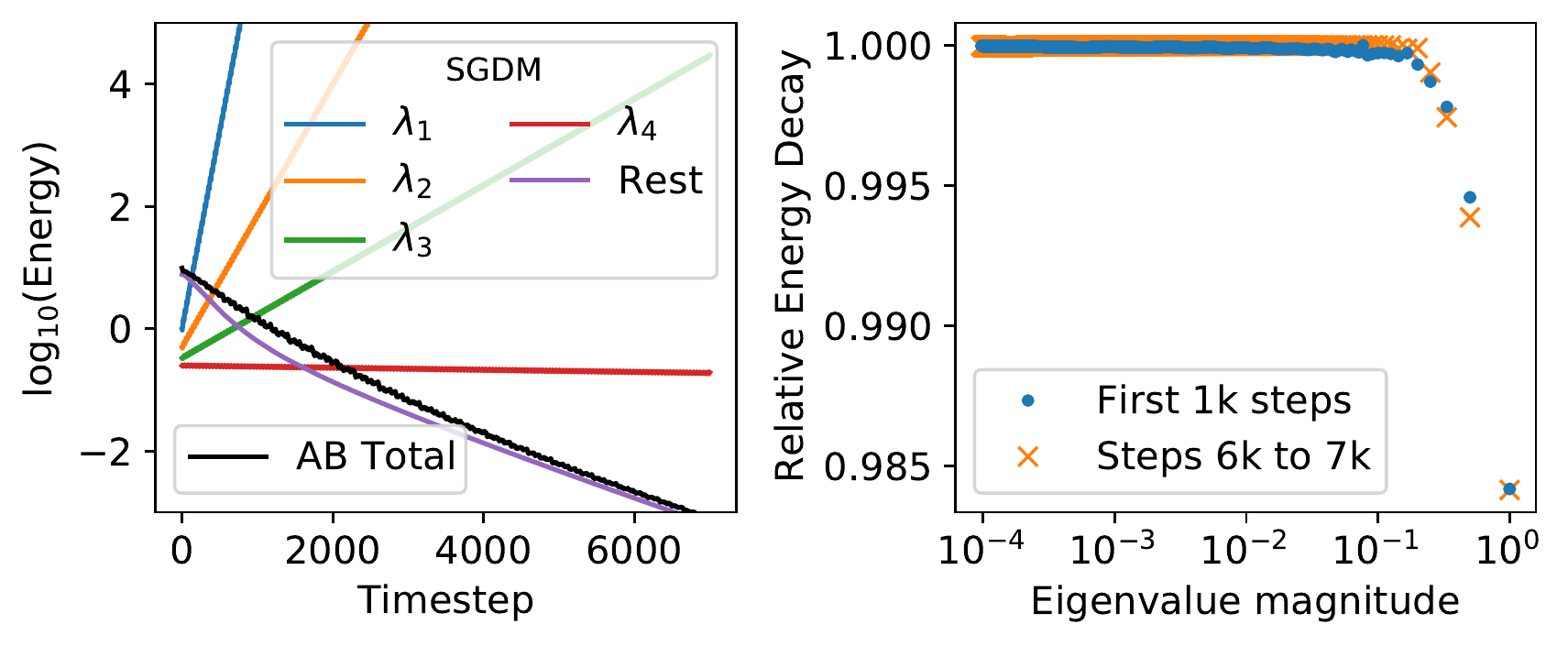}
    \caption{
    AB can stabilize training with gradient delay.
    The learning rate and momentum correspond to values that reach the target loss with AB but are unstable for SGDM in the third row of Figure~\ref{fig:heatmap}. 
    \textbf{Left:} The components corresponding to the three largest eigenvalues are unstable without AB.
    \textbf{Right:} AB dissipates energy in these components on average which stabilizes training.
    }
    \label{fig:cq_dampening_delay}
\end{figure}

The third and forth rows of Figure~\ref{fig:heatmap} show that AB can help with gradient delay. AB can expand the region of convergence and significantly reduce the time required to reach the target loss. Similar to \citeauthor{kosson2020pipelined}~\citeyearpar{kosson2020pipelined}, we note that standard momentum does not seem to help in the delay case but with AB there can be a significant benefit. Delays intuitively cause optimization to overshoot, introducing and amplifying oscillations.
AB seems to help stabilize these oscillations improving convergence in the presence of gradient delay. Figure~\ref{fig:cq_dampening_delay} explores this effect. It shows that AB can dampen high curvature components stabilizing training with gradient delay.

\section{Training Neural Networks}
To measure the effectiveness of AB for training neural networks with gradient delay, we simulate multi-worker ASGD. 
We do this on a single machine by storing a history of the master weights $[\vec{w}_t, \vec{w}_{t-1}, \vec{w}_{t-2} ... \vec{w}_{t-D}]$. 
We then use a chosen algorithm to compute the updated master weights $\vec{w}_{t+1}$ using the delayed gradient  $\vec{g}_{t} = G(\vec{w}_{t-D})$.
In all experiments we use a constant delay $D$, which is representative of an ideal ASGD setting with $D+1$ workers and round robin scheduling. All experiments were implemented using the PyTorch framework \cite{pytorch2019}, and executed on NVIDIA T4 or V100 GPUs.

The main metric we are interested in is the final test accuracy of our trained model compared to a zero-delay, single-worker SGDM baseline. This baseline represents the best possible convergence scenario albeit with no parallelism and no speedup. We evaluate the delay tolerance of algorithms by comparing how much the final test accuracy degrades when training with ASGD and different delays $D$. For consistency, we do not change the per-worker hyperparameters from the original SGDM baseline.
We report experiments on two common image classification tasks: ResNet-20 trained on CIFAR-10 \cite{krizhevsky09learningmultiple} and ResNet-50 trained on ImageNet-1k \cite{krizhevsky2012imagenet}. Hyperparameter settings can be found in Appendix~\ref{hsetting}.

In addition to Adaptive Braking, we also evaluate and compare against a variety of gradient delay mitigation strategies\footnote{Algorithmic details can be found in Appendix~\ref{pseudocode}.}: Shifted Momentum (SM) \cite{giladi2019stabilitysedge}, DANA \cite{hakimi2019dana}, Delay-Compensation (DC) \cite{zheng2017asynchronous}, and  Staleness-Aware (SA) \cite{zhang2016staleness}. 

\subsection{CIFAR-10}
\begin{figure}[h]
    \centering
    \includegraphics[width=\linewidth]{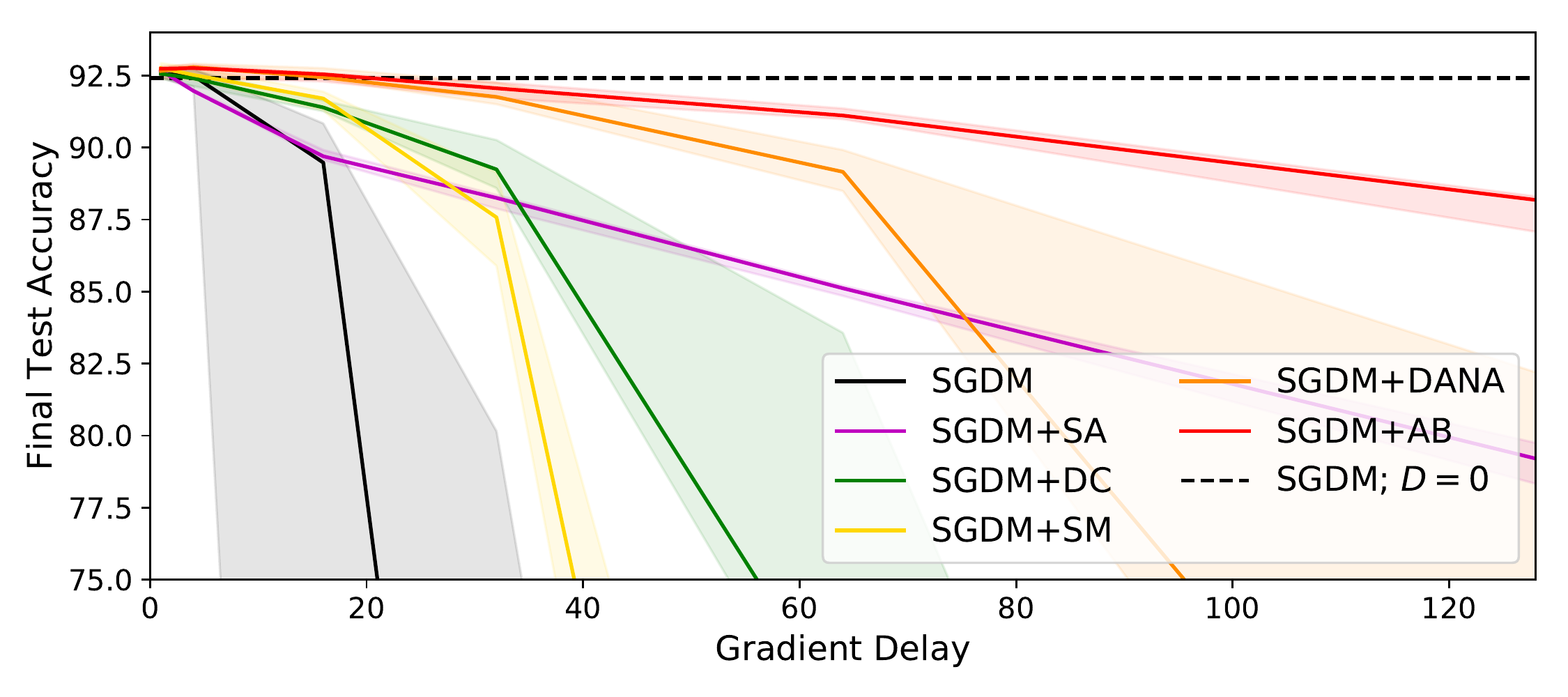}
    \vskip -0.1in
    \caption{ResNet-20 + CIFAR-10 final test accuracy vs delay. AB provides greater delay tolerance than other mitigation strategies. Each line shows the median over five trials.}
    \label{fig: exp2-ab-compare}
\end{figure}

In Figure~\ref{fig: exp2-ab-compare}, we simulate asynchronous training of ResNet-20 on CIFAR-10. We evaluate SGDM combined with other delay mitigation strategies and compare them against SGDM+AB with $\rho=2$ (hyperparameter search shown in Appendix~\ref{sec:cifar10-ext}). 
We find that training with SGDM+AB leads to equivalent accuracy at small-to-moderate delays, and significantly outperforms the other mitigation strategies at large delays ($D = 128$). We also see more stability from run-to-run when compared to the other strategies.

To gain further insight into AB's effects, we measure key metrics $\alpha^i_t$ and $\|\vec{v}^i_t\|$ during CIFAR-10 training and discuss their implications in Appendices \ref{sec:alpha} and \ref{sec:vel-norm}, respectively.

\subsection{ImageNet-1k}

\begin{figure}[h]
    \centering
    \includegraphics[width=\linewidth]{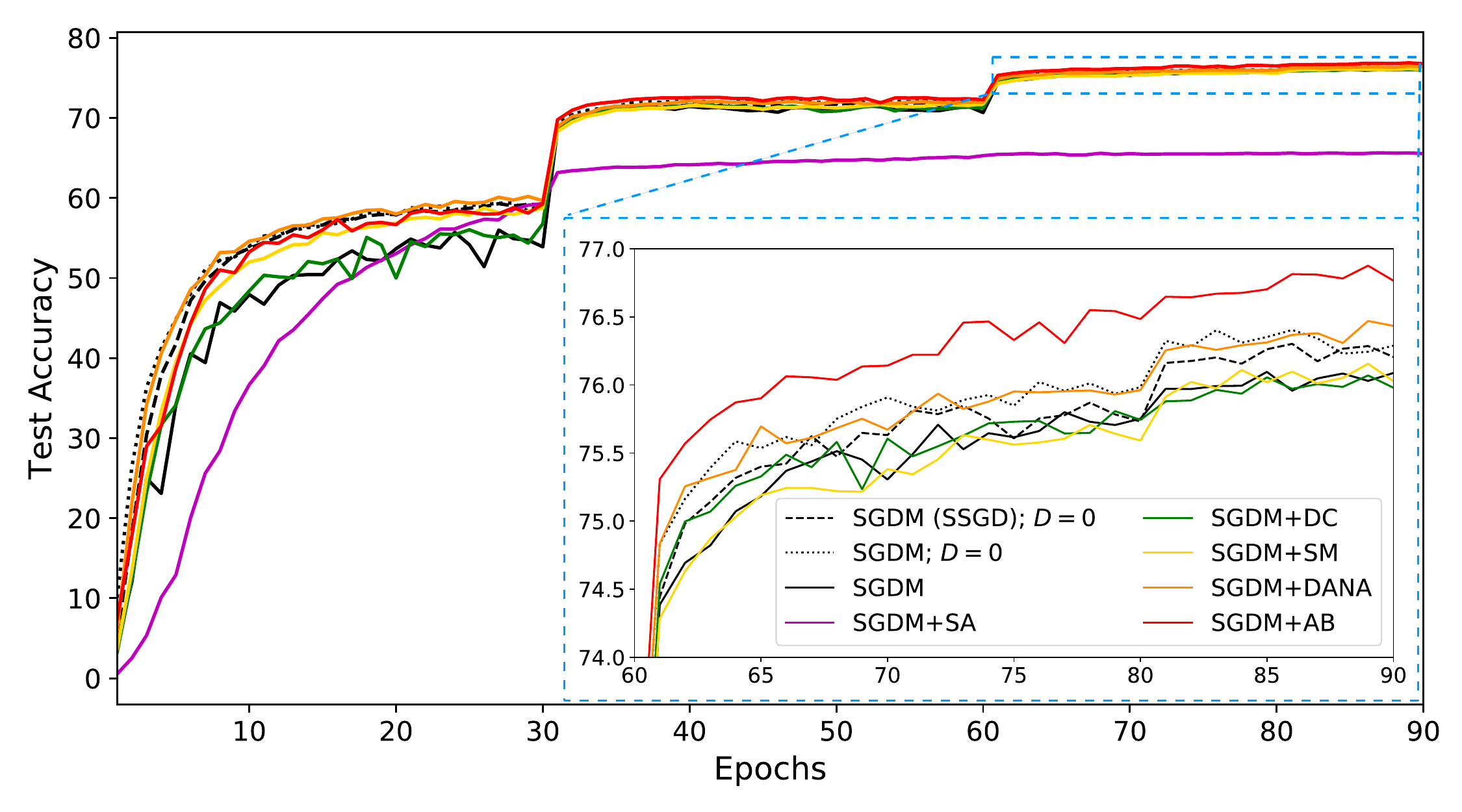}  
    \vskip -0.1in
    \caption{AB outperforms other delay mitigation strategies when training ResNet-50 on ImageNet with a delay of $D=32$.}
    \label{fig:exp3-i1k-compare}
\end{figure}

In Figure~\ref{fig:exp3-i1k-compare}, we simulate asynchronous training of ResNet-50 on ImageNet-1k with a delay of $D = 32$. We compare the vanilla SGDM optimizer to SGDM+AB with $\rho=2$. For our zero-delay baseline, in addition to using a single worker as in the CIFAR-10 experiments, we also include a more realistic Synchronous SGD (SSGD) setup with $D + 1 = 33$ workers. For the SSGD run we use a large batch size of $BS' = 32 * 33 = 1056$ and linearly-scaled learning rate $LR' = 0.00125 * 33 = 0.04125$.

We confirm that training with vanilla SGDM and gradient delay leads to poor convergence at the start of training, and a final test accuracy degradation of -0.24\% compared to the single-worker baseline.
Using SGDM+AB leads to more stable convergence during early training; the test accuracy curve is closer to synchronous training. 
Overall, AB prevents final accuracy degradation for asynchronous training and even outperforms the single-worker baseline by +0.52\%.

We also compare AB with other delay mitigation strategies in the same ASGD setting. We find that SGDM+AB outperforms the other algorithms in terms of final test accuracy. 
Among the other algorithms, SGDM+DANA performs the best, and following a similar trajectory to AB in the early stages of training.
Final test accuracies for all methods are reported in Table~\ref{tab:i1k}.

\begin{table}[t]
    \centering
    \caption{ResNet-50 + ImageNet-1k final test accuracy. The median value over the last 5 epochs is reported.}
    \vskip 0.1in
    \small
    \sc
    \resizebox{\columnwidth}{!}{
    \begin{tabular}{lcccc}
        \toprule
        Algorithm & BS & D & Accuracy & Degradation \\
        \midrule
        SGDM & 32 & 0 & 76.29\% & --- \\
        SGDM (SSGD) & 1056 & 0 & 76.27\% & -0.02\% \\
        \midrule
        SGDM & 32 & 32 & 76.05\% & -0.24\% \\
        SGDM+SA & '' & '' & 65.59\% & -10.70\% \\
        SGDM+DC & '' & '' & 75.99\% & -0.30\% \\
        SGDM+SM & '' & '' & 76.05\% & -0.24\% \\
        SGDM+DANA & '' & '' & 76.38\% & +0.09\% \\
        SGDM+AB & '' & '' & \textbf{76.81\%} & \textbf{+0.52\%} \\
        \bottomrule
    \end{tabular}
    }
    \label{tab:i1k}
\end{table}

\section{Conclusion}
Adaptive Braking scales the gradient based on the alignment of the gradient and velocity. This is a non-linear operation that dampens oscillations along the high-curvature components of the loss surface without affecting the other components much on average. 
It is especially effective in the presence of gradient delay where it can stabilize components that would otherwise be unstable. We show that AB is competitive with state of the art methods for ASGD training.

The increased delay tolerance that AB provides could enable hardware speedups for both data-parallel distributed training as well as pipeline-parallel training with pipelined backpropagation \cite{Ptrowski1993PerformanceAO,chen2012pipelined,Harlap2018PipeDreamFA}.

In this work we have focused on the SGDM optimizer, but future work could propose similar modifications to other optimizers 
such as Adam \cite{kingma2014adam}.

\section*{Acknowledgements}
We thank Joel Hestness, Vithursan Thangarasa, and Xin Wang for for their help and feedback that improved the manuscript.




\bibliography{adabrake}

\begin{thebibliography}{24}
\providecommand{\natexlab}[1]{#1}
\providecommand{\url}[1]{\texttt{#1}}
\expandafter\ifx\csname urlstyle\endcsname\relax
  \providecommand{\doi}[1]{doi: #1}\else
  \providecommand{\doi}{doi: \begingroup \urlstyle{rm}\Url}\fi

\bibitem[Amodei \& Hernandez(2018)Amodei and Hernandez]{amodei2018ai}
Amodei, D. and Hernandez, D.
\newblock {AI} and compute.
\newblock \emph{Heruntergeladen von https://blog. openai. com/aiand-compute},
  2018.

\bibitem[Chen et~al.(2018)Chen, Yang, and Cheng]{chen2018efficient}
Chen, C.-C., Yang, C.-L., and Cheng, H.-Y.
\newblock Efficient and robust parallel dnn training through model parallelism
  on multi-gpu platform.
\newblock \emph{arXiv preprint arXiv:1809.02839}, 2018.

\bibitem[Chen et~al.(2016)Chen, Pan, Monga, Bengio, and
  Jozefowicz]{chen2016revisiting}
Chen, J., Pan, X., Monga, R., Bengio, S., and Jozefowicz, R.
\newblock Revisiting distributed synchronous sgd.
\newblock \emph{arXiv preprint arXiv:1604.00981}, 2016.

\bibitem[Chen et~al.(2012)Chen, Eversole, Li, Yu, and Seide]{chen2012pipelined}
Chen, X., Eversole, A., Li, G., Yu, D., and Seide, F.
\newblock Pipelined back-propagation for context-dependent deep neural
  networks.
\newblock In \emph{Interspeech}. ISCA, September 2012.

\bibitem[Giladi et~al.(2020)Giladi, Nacson, Hoffer, and
  Soudry]{giladi2019stabilitysedge}
Giladi, N., Nacson, M.~S., Hoffer, E., and Soudry, D.
\newblock At stability's edge: How to adjust hyperparameters to preserve minima
  selection in asynchronous training of neural networks?
\newblock In \emph{International Conference on Learning Representations}, 2020.

\bibitem[{Guan} et~al.(2017){Guan}, {Shan}, {Yang}, {Xu}, and
  {Zhang}]{gaun2017delay}
{Guan}, N., {Shan}, L., {Yang}, C., {Xu}, W., and {Zhang}, M.
\newblock Delay compensated asynchronous adam algorithm for deep neural
  networks.
\newblock In \emph{2017 IEEE International Symposium on Parallel and
  Distributed Processing with Applications and 2017 IEEE International
  Conference on Ubiquitous Computing and Communications (ISPA/IUCC)}, pp.\
  852--859, Dec 2017.
\newblock \doi{10.1109/ISPA/IUCC.2017.00130}.

\bibitem[Hakimi et~al.(2019)Hakimi, Barkai, Gabel, and
  Schuster]{hakimi2019dana}
Hakimi, I., Barkai, S., Gabel, M., and Schuster, A.
\newblock {DANA}: Scalable out-of-the-box distributed {ASGD} without retuning,
  2019.

\bibitem[Harlap et~al.(2018)Harlap, Narayanan, Phanishayee, Seshadri, Devanur,
  Ganger, and Gibbons]{Harlap2018PipeDreamFA}
Harlap, A., Narayanan, D., Phanishayee, A., Seshadri, V., Devanur, N.~R.,
  Ganger, G.~R., and Gibbons, P.~B.
\newblock Pipe{D}ream: Fast and efficient pipeline parallel dnn training.
\newblock \emph{ArXiv}, abs/1806.03377, 2018.

\bibitem[Hermans \& Louppe(2018)Hermans and Louppe]{hermans2018gradient}
Hermans, J. and Louppe, G.
\newblock Gradient energy matching for distributed asynchronous gradient
  descent.
\newblock \emph{arXiv preprint arXiv:1805.08469}, 2018.

\bibitem[Kingma \& Ba(2014)Kingma and Ba]{kingma2014adam}
Kingma, D.~P. and Ba, J.
\newblock Adam: A method for stochastic optimization.
\newblock \emph{arXiv preprint arXiv:1412.6980}, 2014.

\bibitem[Kosson et~al.(2020)Kosson, Chiley, Venigalla, Hestness, and
  K{\"o}ster]{kosson2020pipelined}
Kosson, A., Chiley, V., Venigalla, A., Hestness, J., and K{\"o}ster, U.
\newblock Pipelined backpropagation at scale: Training large models without
  batches.
\newblock \emph{arXiv preprint arXiv:2003.11666}, 2020.

\bibitem[Krizhevsky(2009)]{krizhevsky09learningmultiple}
Krizhevsky, A.
\newblock Learning multiple layers of features from tiny images.
\newblock Technical report, 2009.

\bibitem[Krizhevsky et~al.(2012)Krizhevsky, Sutskever, and
  Hinton]{krizhevsky2012imagenet}
Krizhevsky, A., Sutskever, I., and Hinton, G.~E.
\newblock Imagenet classification with deep convolutional neural networks.
\newblock In Pereira, F., Burges, C. J.~C., Bottou, L., and Weinberger, K.~Q.
  (eds.), \emph{Advances in Neural Information Processing Systems 25}, pp.\
  1097--1105. Curran Associates, Inc., 2012.

\bibitem[Lian et~al.(2015)Lian, Huang, Li, and Liu]{lian2015asynchronous}
Lian, X., Huang, Y., Li, Y., and Liu, J.
\newblock Asynchronous parallel stochastic gradient for nonconvex optimization.
\newblock In Cortes, C., Lawrence, N.~D., Lee, D.~D., Sugiyama, M., and
  Garnett, R. (eds.), \emph{Advances in Neural Information Processing Systems
  28}, pp.\  2737--2745. Curran Associates, Inc., 2015.

\bibitem[Mitliagkas et~al.(2016)Mitliagkas, Zhang, Hadjis, and
  R{\'e}]{mitliagkas2016asynchrony}
Mitliagkas, I., Zhang, C., Hadjis, S., and R{\'e}, C.
\newblock Asynchrony begets momentum, with an application to deep learning.
\newblock In \emph{2016 54th Annual Allerton Conference on Communication,
  Control, and Computing (Allerton)}, pp.\  997--1004. IEEE, 2016.

\bibitem[O’Donoghue \& Candes(2012)O’Donoghue and
  Candes]{odonoghue3982adaptive}
O’Donoghue, B. and Candes, E.
\newblock Adaptive restart for accelerated gradient schemes.
\newblock \emph{arXiv preprint arXiv:1204.3982}, 2012.

\bibitem[Paszke et~al.(2019)Paszke, Gross, Massa, Lerer, Bradbury, Chanan,
  Killeen, Lin, Gimelshein, Antiga, Desmaison, Kopf, Yang, DeVito, Raison,
  Tejani, Chilamkurthy, Steiner, Fang, Bai, and Chintala]{pytorch2019}
Paszke, A., Gross, S., Massa, F., Lerer, A., Bradbury, J., Chanan, G., Killeen,
  T., Lin, Z., Gimelshein, N., Antiga, L., Desmaison, A., Kopf, A., Yang, E.,
  DeVito, Z., Raison, M., Tejani, A., Chilamkurthy, S., Steiner, B., Fang, L.,
  Bai, J., and Chintala, S.
\newblock Pytorch: An imperative style, high-performance deep learning library.
\newblock In \emph{Advances in Neural Information Processing Systems 32}, pp.\
  8024--8035. Curran Associates, Inc., 2019.

\bibitem[P{\'e}trowski et~al.(1993)P{\'e}trowski, Dreyfus, and
  Girault]{Ptrowski1993PerformanceAO}
P{\'e}trowski, A., Dreyfus, G., and Girault, C.
\newblock Performance analysis of a pipelined backpropagation parallel
  algorithm.
\newblock \emph{IEEE transactions on neural networks}, 4 6:\penalty0 970--81,
  1993.

\bibitem[{Rigazzi}(2019)]{rigazzi2019dc3sgd}
{Rigazzi}, A.
\newblock Dc-s3gd: Delay-compensated stale-synchronous sgd for large-scale
  decentralized neural network training.
\newblock In \emph{2019 IEEE/ACM Third Workshop on Deep Learning on
  Supercomputers (DLS)}, pp.\  62--68, 2019.

\bibitem[Shallue et~al.(2019)Shallue, Lee, Antognini, Sohl-Dickstein, Frostig,
  and Dahl]{shallue2019measuring}
Shallue, C.~J., Lee, J., Antognini, J., Sohl-Dickstein, J., Frostig, R., and
  Dahl, G.~E.
\newblock Measuring the effects of data parallelism on neural network training.
\newblock \emph{Journal of Machine Learning Research}, 20\penalty0
  (112):\penalty0 1--49, 2019.

\bibitem[Yang et~al.(2019)Yang, Zhang, Li, R{\'e}, Aberger, and
  De~Sa]{yang2019pipemare}
Yang, B., Zhang, J., Li, J., R{\'e}, C., Aberger, C.~R., and De~Sa, C.
\newblock Pipemare: Asynchronous pipeline parallel dnn training.
\newblock \emph{arXiv preprint arXiv:1910.05124}, 2019.

\bibitem[Zhang et~al.(2019)Zhang, Li, Nado, Martens, Sachdeva, Dahl, Shallue,
  and Grosse]{zhang2019algorithmic}
Zhang, G., Li, L., Nado, Z., Martens, J., Sachdeva, S., Dahl, G., Shallue, C.,
  and Grosse, R.~B.
\newblock Which algorithmic choices matter at which batch sizes? insights from
  a noisy quadratic model.
\newblock In \emph{Advances in Neural Information Processing Systems}, pp.\
  8194--8205, 2019.

\bibitem[Zhang et~al.(2016)Zhang, Gupta, Lian, and Liu]{zhang2016staleness}
Zhang, W., Gupta, S., Lian, X., and Liu, J.
\newblock Staleness-aware async-sgd for distributed deep learning.
\newblock In \emph{Proceedings of the Twenty-Fifth International Joint
  Conference on Artificial Intelligence}, IJCAI’16, pp.\  2350–2356. AAAI
  Press, 2016.
\newblock ISBN 9781577357704.

\bibitem[Zheng et~al.(2017)Zheng, Meng, Wang, Chen, Yu, Ma, and
  Liu]{zheng2017asynchronous}
Zheng, S., Meng, Q., Wang, T., Chen, W., Yu, N., Ma, Z.-M., and Liu, T.-Y.
\newblock Asynchronous stochastic gradient descent with delay compensation.
\newblock In \emph{Proceedings of the 34th International Conference on Machine
  Learning - Volume 70}, ICML'17, pp.\  4120--4129. JMLR.org, 2017.

\end{thebibliography}
\bibliographystyle{icml2020}

\clearpage
\pagebreak
\appendix

\section{Related Work}
Asynchronous methods are used to improve compute utilization for neural network training but introduce gradient staleness.
Gradients are stale because the gradient is computed using weight from $D$ time steps ago, $\vec{g}_t = G\left( \vec{w}_{t-D} \right)$.
To mitigate this \citeauthor{giladi2019stabilitysedge}~\citeyearpar{giladi2019stabilitysedge} propose adding delay to the velocity as well, $\vec{v}_{t-D}$.
They do this by tracking an independent velocity for each worker and updating the master weights using the current worker's velocity.
Another class of mitigation strategies attempts to predict future weights $\hat{\vec{w}}_{t-D} \approx \vec{w}_t$ for use in the gradient computation, $\vec{g}_t = G\left( \hat{\vec{w}}_{t-D} \right)$.
Most methods \cite{chen2018efficient,hakimi2019dana,kosson2020pipelined} use the velocity vector to estimate the future weights.

\citeauthor{zhang2016staleness}~\citeyearpar{zhang2016staleness} propose
Staleness-Aware (SA) and show that down-weighing the gradients based on the delay $D$ (gradient penalization) can improve asynchronous training.
\citeauthor{kosson2020pipelined}~\citeyearpar{kosson2020pipelined} characterize the impulse response of gradients in the optimization process and modify the delayed impulse response to match the non-delayed setting in a technique called Spike Compensation.

Delay Compensated ASGD \cite{zheng2017asynchronous} and its variants \cite{gaun2017delay,rigazzi2019dc3sgd} estimate the gradient using the first two terms of the Taylor expansion of the delayed gradient function.
Using the Taylor expansion of the delayed gradient function requires estimating the Hessian and storing the old weights. Applying DC-ASGD with a velocity approximation for the weight change is closely related to element-wise Adaptive Braking (See Appendix \ref{sec:ab-dc}). 

\citeauthor{odonoghue3982adaptive}~\citeyearpar{odonoghue3982adaptive} use a method called Adaptive Restart (AR) to dampen oscillations and speed up optimization. Adaptive Restart resets the velocity, $\vec{v} = \vec{0}$, when $\vec{g}^T \cdot \vec{v} < 0$ which can be viewed as a measure of alignment. AB also measures alignment using cosine similarity but applies a continuous correction to $\vec{g}$ rather than a discrete reset of $\vec{v}$. This makes AB more applicable in a noisy optimization setting such as SGD. Periodically resetting the step direction is also used in nonlinear conjugate gradient optimization methods. Adaptive Braking can be seen as a form of nonlinear conjugate gradient optimization since the step direction accumulation is adaptively adjusted based on the current gradient. There are many variations of nonlinear conjugate gradient optimization but to the best of our knowledge, none of these forms are exactly equivalent to Adaptive Braking.

The rest of this section shows the algorithmic details of the methods we compare against in our experiments.

\label{pseudocode}
\subsection{Asynchronous SGD (ASGD)}

\begin{algorithm}[H]
	\caption{Momentum-ASGD: worker $j$}
	\label{alg:ASGD_worker}
	\centering
	\begin{algorithmic}
	    \setlength{\itemindent}{-0.5em} 
	    \STATE Always do:
		\STATE \quad Receive parameters $\vec{w}_{t-D}$ from the master
		\STATE \quad Compute gradient: $\vec{g}_{t;j} = G(\vec{w}_{t-D})$
		\STATE \quad Send $\vec{g}_{t;j}$ to the master
	\end{algorithmic}
\end{algorithm}

\begin{algorithm}[H]
\caption{Momentum-ASGD: master}
\label{alg:ASGD_master}
\begin{algorithmic}
    \setlength{\itemindent}{-0.5em} 
    \STATE For t = 1...T do:
    \STATE \quad Receive gradient $\vec{g}_{t;j}$ from worker $j$
    \STATE \quad Update momentum: $\vec{v}_{t+1} = m \vec{v}_t+\vec{g}_{t;j}$
    \STATE \quad Update master's weights: $\vec{w}_{t+1} = \vec{w}_t-\eta_t \vec{v}_{t+1}$
    \STATE \quad Send $\vec{w}_{t+1}$ to worker $j$
\end{algorithmic}
\end{algorithm}

\subsection{Staleness-Aware}

Staleness-Aware divides the original learning rate by the delay of the current gradient in each update step.

\begin{algorithm}[H]
\caption{Staleness-Aware: master}
\label{alg:Stale-Aware_master}
\begin{algorithmic}
    \setlength{\itemindent}{-0.5em} 
    \STATE Initialize an iteration array: $iter = [0]*N$
    \STATE For t = 1...T do:
    \STATE \quad Receive gradient $\vec{g}_{t;j}$ from worker $j$
    \STATE \quad Calculate worker $j$'s delay: $D_t = t-iter[j]$
    \STATE \quad Update momentum: $\vec{v}_{t+1} = m \vec{v}_t+\vec{g}_{t;j}$
    \STATE \quad Update master: $\vec{w}_{t+1} = \vec{w}_t-\frac{\eta_t}{D_t} \vec{v}_{t+1}$
    \STATE \quad Send $\vec{w}_{t+1}$ to worker $j$
    \STATE \quad Save current iteration: $iter[j] = t$
\end{algorithmic}
\end{algorithm}

\subsection{Shifted Momentum}
Shifted Momentum assigns an independent velocity $\vec{v}_{t;j}$ to each worker $j$, and updates the master weights using the current worker's velocity.

\begin{algorithm}[H]
	\caption{Shifted Momentum: worker $j$}
	\label{alg:Shifted_worker}
	\centering
	\begin{algorithmic}
	    \setlength{\itemindent}{-0.5em} 
	    \STATE Always do:
		\STATE \quad Receive parameters $\vec{w}_{t-D}$ from the master
		\STATE \quad Compute gradient: $\vec{g}_{t;j} = G(\vec{w}_{t-D})$
		\STATE \quad Update momentum $\vec{v}_{t+1;j} = m \vec{v}_{t;j}+\vec{g}_{t;j}$
		\STATE \quad Send $\vec{v}_{t+1;j}$ to the master
	\end{algorithmic}
\end{algorithm}

\begin{algorithm}[H]
\caption{Shifted Momentum: master}
\label{alg:Shifted_master}
\begin{algorithmic}
    \setlength{\itemindent}{-0.5em} 
    \STATE For t = 1...T do:
    \STATE \quad Receive gradient $\vec{v}_{t+1;j}$ from worker $j$
    \STATE \quad Update master's weights: $\vec{w}_{t+1} = \vec{w}_t - \eta_t \vec{v}_{t+1;j}$
    \STATE \quad Send $\vec{w}_{t+1}$ to worker $j$
\end{algorithmic}
\end{algorithm}

\subsection{DANA}
DANA assigns an independent velocity $v_{t;j}$ to each worker $j$, and computes the gradient on estimated future weights.

\begin{algorithm}[H]
	\caption{DANA: worker $j$}
	\label{alg:dana_worker}
	\centering
	\begin{algorithmic}
	    \setlength{\itemindent}{-0.5em} 
	    \STATE Always do:
		\STATE \quad Receive parameters $\hat{\vec{w}}_{t-D}$ from the master
		\STATE \quad Compute gradient: $\vec{g}_{t;j} = G(\vec{w}_{t-D})$
		\STATE \quad Update momentum: $\vec{v}_{t+1;j} = m \vec{v}_{t;j}+\vec{g}_{t;j}$
		\STATE \quad Send $\vec{v}_{t+1;j}$ to the master
	\end{algorithmic}
\end{algorithm}

\begin{algorithm}[H]
\caption{DANA: master}
\label{alg:dana_master}
\begin{algorithmic}
    \setlength{\itemindent}{-0.5em} 
    \STATE For t = 1...T do:
    \STATE \quad Receive gradient $\vec{v}_{t+1;j}$ from worker $j$
    \STATE \quad Update master's weights: $\vec{w}_{t+1} = \vec{w}_t-\eta_t \vec{v}_{t+1;j}$
    \STATE \quad Estimates future weights: $\hat{\vec{w}}_{t+1} = \vec{w}_t-\eta_t m \sum_j \vec{v}_{t+1;j}$
    \STATE \quad Send $\hat{\vec{w}}_{t+1}$ to worker $j$
\end{algorithmic}
\end{algorithm}

\subsection{Delay-Compensated ASGD}
Delay-Compensated ASGD approximates the Hessian of the loss surface and corrects the delayed gradient based on the weight inconsistency.
\begin{algorithm}[H]
\caption{Delay-Compensated ASGD: master}
\label{alg:DC-ASGD_master}
\begin{algorithmic}
    \setlength{\itemindent}{-0.5em} 
    \STATE For t = 1...T do:
    \STATE \quad Receive gradient $\vec{g}_{t;j}$ from worker $j$
    \STATE \quad Compensate gradient: $\hat{\vec{g}}_{t;j} = \vec{g}_{t;j} + \nabla \vec{g}_{t;j} \cdot (\vec{w}_{t} - \vec{w}_{t-D}) $ 
    \STATE \quad Update momentum: $\vec{v}_{t+1} = m \vec{v}_t+\hat{\vec{g}}_{t;j}$
    \STATE \quad Update master's weights: $\vec{w}_{t+1} = \vec{w}_t-\eta_t  \vec{v}_{t+1}$
    \STATE \quad Send $\vec{w}_{t+1}$ to worker $j$
\end{algorithmic}
\end{algorithm}

where $\nabla \vec{g}_{t;j}$ is approximated with $\lambda_t \cdot \text{diag}(\vec{g}_{t;j} \odot  \vec{g}_{t;j})$ and $\lambda_t$ is the variance control parameter, set using a moving-average as described in the original paper. We note that this algorithm is modified to work with SGDM.

\subsection{Adaptive Braking}

\begin{algorithm}[H]
\caption{Adaptive Braking: master}
\label{alg:AB_master}
\begin{algorithmic}
    \setlength{\itemindent}{-0.5em} 
    \STATE For t = 1...T do:
    \STATE \quad Receive gradient $\vec{g}_{t;j}$ from worker $j$
    \STATE \quad For i in parameter groups do:
    \STATE \quad \quad Compute braking: $\alpha^i_{t} = 1 - \rho \cos \angle \left( \vec{g}^i_{t}, \vec{v}^i_t \right)$
    \STATE \quad \quad Update momentum: $\vec{v}^i_{t+1} = m \vec{v}^i_t+\alpha^i_{t}\vec{g}^i_{t;j}$
    \STATE \quad \quad Update master's weights: $\vec{w}^i_{t+1} = \vec{w}^i_t-\eta_t  \vec{v}^i_{t+1}$
    \STATE \quad Send $\vec{w}_{t+1}$ to worker $j$
\end{algorithmic}
\end{algorithm}

\section{Hyperparameter Settings}
\label{hsetting}

The per-worker hyperparameter settings used in our neural network training experiments are listed in Table~\ref{tab:hyperparameters}. 
For CIFAR-10, we choose to use a small batch size of 32 rather than the standard setting of 128 to showcase a training setup with high momentum, which is where Adaptive Braking is most effective. 
For ImageNet-1k we use a per-worker batch size of 32 to reflect a common SSGD training setup with 8 GPUs and a total batch size of 256, and choose a momentum of 0.99 based on hyperparameter searches performed by \citeauthor{shallue2019measuring}~\citeyearpar{shallue2019measuring}.
For DC we use the adaptive form of the algorithm and adopt the original paper's hyperparameters. The other mitigation strategies are hyperparameter-free. 

\begin{table}[h]
    \centering
    \vskip -0.1in
    \caption{Single-worker hyperparameter settings used for ASGD experiments.}
    \vskip 0.1in
    \small
    \sc
    \resizebox{\columnwidth}{!}{
    \begin{tabular}{lcc}
        \toprule
        Parameter & CIFAR-10 & ImageNet1k \\
        \midrule
        Model Architecture & ResNet-20 & ResNet-50 \\
        Per-Worker Batch size & 32 & 32 \\
        Initial learning rate ($\eta$) & 0.01 & 0.0125 \\
        Momentum ($m$) & 0.95 & 0.99 \\
        Weight decay ({$\lambda$}) & 5e-4 & 1e-4 \\
        Epochs & [100, 50, 50, 50] & [30, 30, 20, 10] \\
        LR decay & 0.1 & 0.1 \\
        LR warmup epochs  & 0 & 5 \\
        \bottomrule
    \end{tabular}
    }
    \label{tab:hyperparameters}
\end{table}

\section{CIFAR-10 Extended Results}
\label{sec:cifar10-ext}

\begin{table*}[t]
\centering
\caption{ResNet-20 + CIFAR-10 final test accuracy, trained with different delays $D$. The training hyperparameters are listed in Table \ref{tab:hyperparameters}. Each reported accuracy is a median over 5 trials. For each trial, we use the median test accuracy over the last 10 epochs of training.}
\vskip 0.1in
\small
\sc
\begin{tabular}{lccccccc}
\toprule
Algorithm & D=0 & D=1 & D=4 & D=16 & D=32 & D=64 & D=128 \\
\midrule
SGDM & 92.41\% & 92.34\% & 92.16\% & 90.41\% & 84.03\% & 10.09\% & 10.00\% \\
SGDM+AB, $\rho=0.5$ & 92.36\% & \textbf{92.61\%} & 92.16\% & 91.78\% & 89.22\% & 25.25\% & 45.11\% \\
SGDM+AB, $\rho=1$ & \textbf{92.61\%} & 92.51\% & 92.39\% & 92.18\% & 91.67\% & 89.82\% & 84.15\% \\
SGDM+AB, $\rho=2$ & 92.47\% & 92.43\% & 92.46\% & \textbf{92.27\%} & \textbf{91.99\%} & \textbf{91.21\%} & 89.98\% \\
SGDM+AB, $\rho=3$ & 92.44\% & 92.44\% & \textbf{92.54\%} & 91.87\% & 91.82\% & 90.69\% & 88.22\% \\
SGDM+AB, $\rho=4$ & 92.12\% & 91.98\% & 92.07\% & 91.57\% & 91.50\% & 90.87\% & \textbf{90.07\%} \\
SGDM+AB, $\rho=5$ & 92.12\% & 91.94\% & 92.01\% & 90.97\% & 90.41\% & 89.35\% & 87.06\% \\
\bottomrule
\end{tabular}

\label{tab:cifar10-ext}
\end{table*}

\begin{figure}[t]
    \centering
    \includegraphics[width=\linewidth]{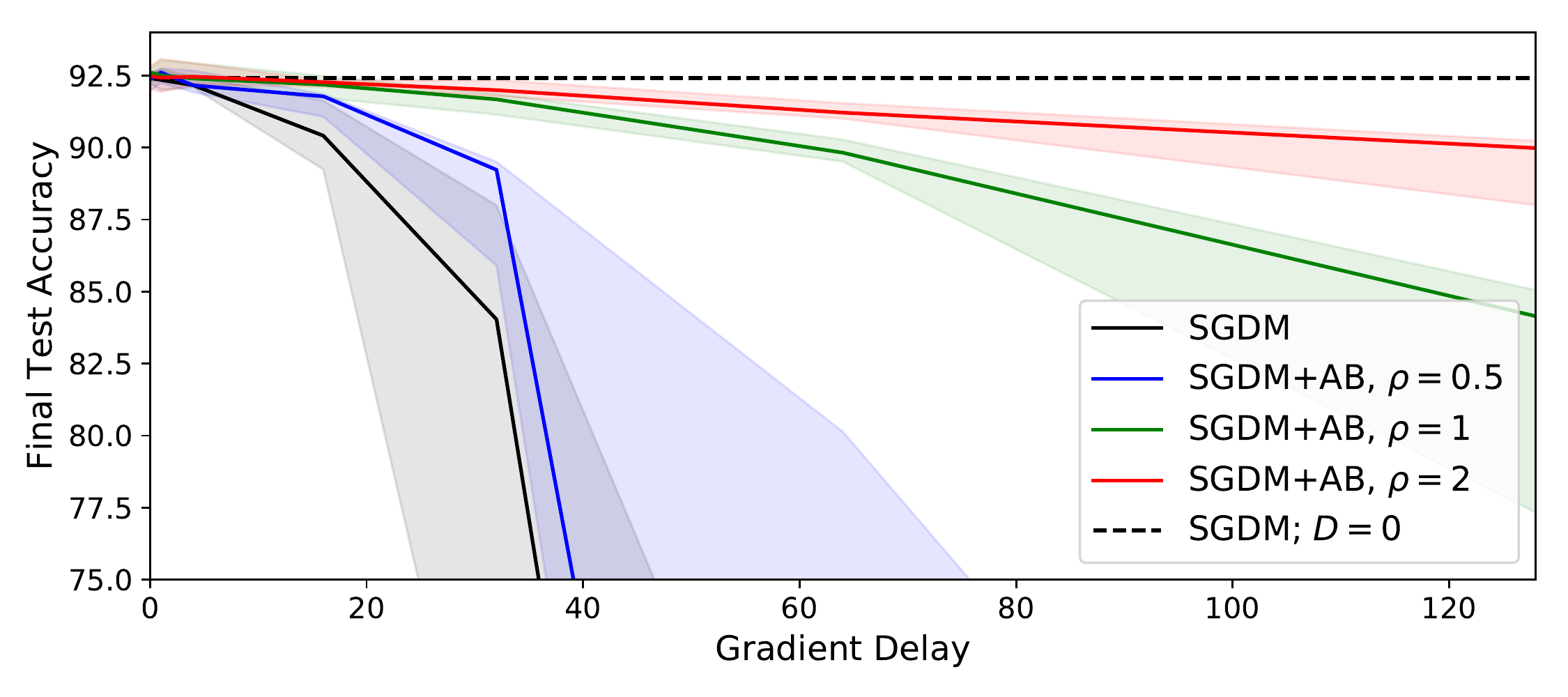}
    \vskip -0.1in
    \caption{ResNet-20 + CIFAR-10 final test accuracy for different delays. Adaptive Braking improves the delay tolerance of SGDM when training in an ASGD setting.}
    \label{fig: exp1-delay-tolerance}
\end{figure}

In Figure~\ref{fig: exp1-delay-tolerance}, we simulate asynchronous training of ResNet-20 on CIFAR-10 and measure the delay tolerance of SGDM with or without Adaptive Braking. Each experiment is repeated 5 times and the median final test accuracy is plotted. 

We find that AB greatly improves the delay tolerance of SGDM. In particular, we can train asynchronously with a gradient delay of $D=32$ with only a -0.42\% drop in test accuracy. Even at extreme settings with $D=128$, the degradation is only -2.43\%, while vanilla SGDM fails to converge at all. We also find that the delay tolerance improves as $\rho$ is increased from $0.5$ to $2.0$. 

In Table~\ref{tab:cifar10-ext} we list extended results with more settings of braking coefficient $\rho$. The results suggest that larger $\rho$ should be used for larger delays. The choice of $\rho = 2$ is the most consistent across $D = [0, 1, 4, 16, 32, 64, 128]$, performing best or second-best in almost all delay settings.

\section{Parameter Grouping for AB}
\label{sec:param-grouping}
The AB gradient scaling factor $\alpha^i_t$ is non-linear with respect to $\vec{g}^i_t$ and $\vec{v}^i_t$, and depends on the granularity with which the model parameters are grouped. We consider three levels of granularity for grouping:
\begin{itemize}
    \item \textbf{Per tensor:} This is based on the default grouping of parameters into tensors in PyTorch. In this case each convolutional or linear layer has a weight tensor which contains all the multiplicative weights and optionally a bias which is a separate tensor. Normalization layers have their own bias and scaling tensors.
    \item \textbf{Per filter:} Here the weights of each neuron or filter are treated separately. The biases and other parameters such as those in the normalization layers are still grouped per tensor.
    \item \textbf{Per element:} Here each parameter is treated separately, and the scaling coefficient reduces to $\alpha_t^i = 1 - \rho \text{ sgn}(g_t^i \cdot v_t^i)$.
\end{itemize}

In Table \ref{tab:grouping} we find that using a filter-wise grouping of parameters leads to the best performance for ResNet-20 trained on CIFAR-10, even when accounting for different optimal settings of $\rho$ for each grouping method. Therefore we use filter-wise Adaptive Braking for all of our experiments.

\begin{table}[b!]
\centering
\caption{ResNet-20 + CIFAR-10 final test accuracy, trained with a delay of $D=32$. Filter-wise grouping outperforms tensor- or element-wise grouping. The optimal setting of $\rho$ for each method is highlighted.}
\vskip 0.1in
\small
\sc
\resizebox{\columnwidth}{!}{%
\begin{tabular}{lccc}
\toprule
Algorithm & Tensor & Filter & Element \\
\midrule
SGDM+AB, $\rho=0.25$ & 88.70\% & 85.46\% & 88.20\% \\
SGDM+AB, $\rho=0.5$ & 87.50\% & 89.24\% & \textbf{90.35\%} \\
SGDM+AB, $\rho=1$ & 91.38\% & 91.67\% & 89.54\% \\
SGDM+AB, $\rho=2$ & \textbf{91.71\%} & \textbf{92.14\%} & 88.97\% \\
SGDM+AB, $\rho=4$ & 91.43\% & 91.15\% & 88.22\% \\
\bottomrule
\end{tabular}
}
\label{tab:grouping}
\end{table}

\section{AB with Weight Decay} \label{sec:ab-weight-decay}
When weight decay is used with Adaptive Braking, we add the weight decay term to the velocity independently, and do not consider the weight decay to be part of the gradient when computing the cosine similarity:

\begin{align}
    \alpha^i_{t} &= 1 - \rho \frac{\langle \vec{g}^i_{t}, \vec{v}^i_t \rangle}{\max(\|\vec{g}^i_{t}\| \|\vec{v}^i_t\|, \epsilon)} \\
    &\approx 1 - \rho \cos \angle \left( \vec{g}^i_{t}, \vec{v}^i_t \right) \nonumber \\    
    \vec{v}^i_{t+1}&= m \vec{v}^i_t + \alpha^i_t \vec{g}^i_{t} + \lambda \vec{w}^i_t \\
    \vec{w}^i_{t+1} &= \vec{w}^i_t - \eta \vec{v}^i_{t+1}
\end{align}

This helps prevent $\alpha^i_t$ from being skewed by the weight decay term, which is correlated across steps.

\section{AB compared with DC}
\label{sec:ab-dc}

Under a particular approximation, delay-compensated ASGD has a similar form to Adaptive Braking. DC attempts to correct the delayed gradient $\vec{g_t}$ by measuring the change in the master weights, and using a Hessian approximation to estimate the up-to-date gradient $\vec{\hat{g}_t}$ at the current master weights:

\begin{align}
    \vec{g}_t &= G(\vec{w}_{t-D}) \\
    \vec{\hat{g}}_t &=  G(\vec{w}_t) \approx \vec{g}_t + \lambda \vec{H}_t \cdot (\vec{w}_t - \vec{w})_{t-D}  \label{eq:delay_gd_2ord}\\
     &\approx \vec{g}_t + \lambda (\vec{g}_t \cdot \vec{g}_t^T) \cdot (\vec{w}_t - \vec{w}_{t-D}) \label{eq:delay_gd_2ord_approx_H} \\
     &\approx \vec{g}_t + \lambda \cdot \text{diag}(\vec{g}_t \odot \vec{g}_t) \cdot (\vec{w}_t - \vec{w}_{t-D}) \label{eq:dc_diag}\\
     &= \vec{g}_t + \lambda (\vec{g}_t \odot \vec{g}_t) \odot (\vec{w}_t - \vec{w}_{t-D}) \label{eq:dc}
\end{align}

\citeauthor{zheng2017asynchronous}~\citeyearpar{zheng2017asynchronous} use the Taylor Series expansion of the delayed gradient but truncate higher order terms \eqref{eq:delay_gd_2ord}. $\vec{H}_t$ is the hessian at time $t$, and $\lambda$ is a hyperparameter used to control the strength of the second-order correction. They then approximate $\vec{H}_t \approx \vec{g}_t \cdot  \vec{g}_t^T \approx  \text{diag}(\vec{g}_t \odot \vec{g}_t)$ which makes the compensation method an element-wise operation.
To arrive at AB, we maintain the outer-product form \eqref{eq:delay_gd_2ord_approx_H} for the remainder of this section.

During SGDM training, the weight update at each step $t$ is $-\eta \vec{v_t}$. If we assume the velocity over the last $D$ steps is relatively unchanged, then we can approximate the total weight change from step $(t-D)$ to step $t$ as:

\begin{align}
    \vec{w}_t - \vec{w}_{t-D} &= -\sum_{i=0}^{D-1} \eta \vec{v}_{t-i} \\
     &\approx -\sum_{i=0}^{D-1} \eta \vec{v}_t \\
     &= -\eta D \vec{v}_t
\end{align}

Substituting this approximation into \eqref{eq:delay_gd_2ord_approx_H}, we end up with a gradient scaling term that involves a dot product of the gradient and the velocity:

\begin{align}
    \hat{\vec{g}}_t &\approx \vec{g}_t + \lambda (\vec{g}_t \cdot \vec{g}_t^T) \cdot (-\eta D \vec{v}_t)  \\
     &= \vec{g}_t (1 -  \lambda \eta D (\vec{g}_t^T \cdot \vec{v}_t)) \\
     &= \vec{g}_t (1 -  \lambda' (\vec{g}_t^T \cdot \vec{v}_t))
\end{align}

Finally, we can use an adaptive setting of $\lambda'$, normalizing by the magnitudes of the gradient and velocity at time $t$. At this point we are no longer approximating the master weight gradient $\vec{\hat{g}_t}$, so we adjust notation:

\begin{align}
    \lambda'_t &= \frac{\lambda'_0}{\|\vec{g}_t\| \cdot \|\vec{v}_t\|} \\
    \vec{g}'_t &= \vec{g}_t (1 -  \lambda'_0  \cos \angle (\vec{g}_t, \vec{v}_t)) \label{eq:dc-approx}
\end{align}

The gradient scaling term in \eqref{eq:dc-approx} is now exactly $\alpha_t = 1 - \rho \cos \angle (\vec{g}_t, \vec{v}_t)$ used for AB. Note that for AB the braking coefficient $\rho$ is chosen independently rather than set based on the learning rate $\eta$ and delay $D$.

\section{Gradient scale $\alpha$ and braking coefficient $\rho$}
\label{sec:alpha}

The strength of AB's gradient scaling $\alpha^i_t$ depends on the choice of braking coefficient $\rho$. In general, the optimal setting of $\rho$ is task-dependent and can be optimized as a hyperparameter, but we find that values in $\rho \in [0.5, 2]$ work well across different delays and model architectures. Note that if $\rho$ is set larger than 1, it is possible for the gradient scaling at a particular step to be negative, but we find this to rarely happen in practice. We have also experimented with clamping $\alpha^i_t$ to be non-negative but do not see a significant effect on convergence, so for simplicity we do not perform clamping in the standard form of AB.

\begin{figure}[t]
\centering
\includegraphics[width=\linewidth]{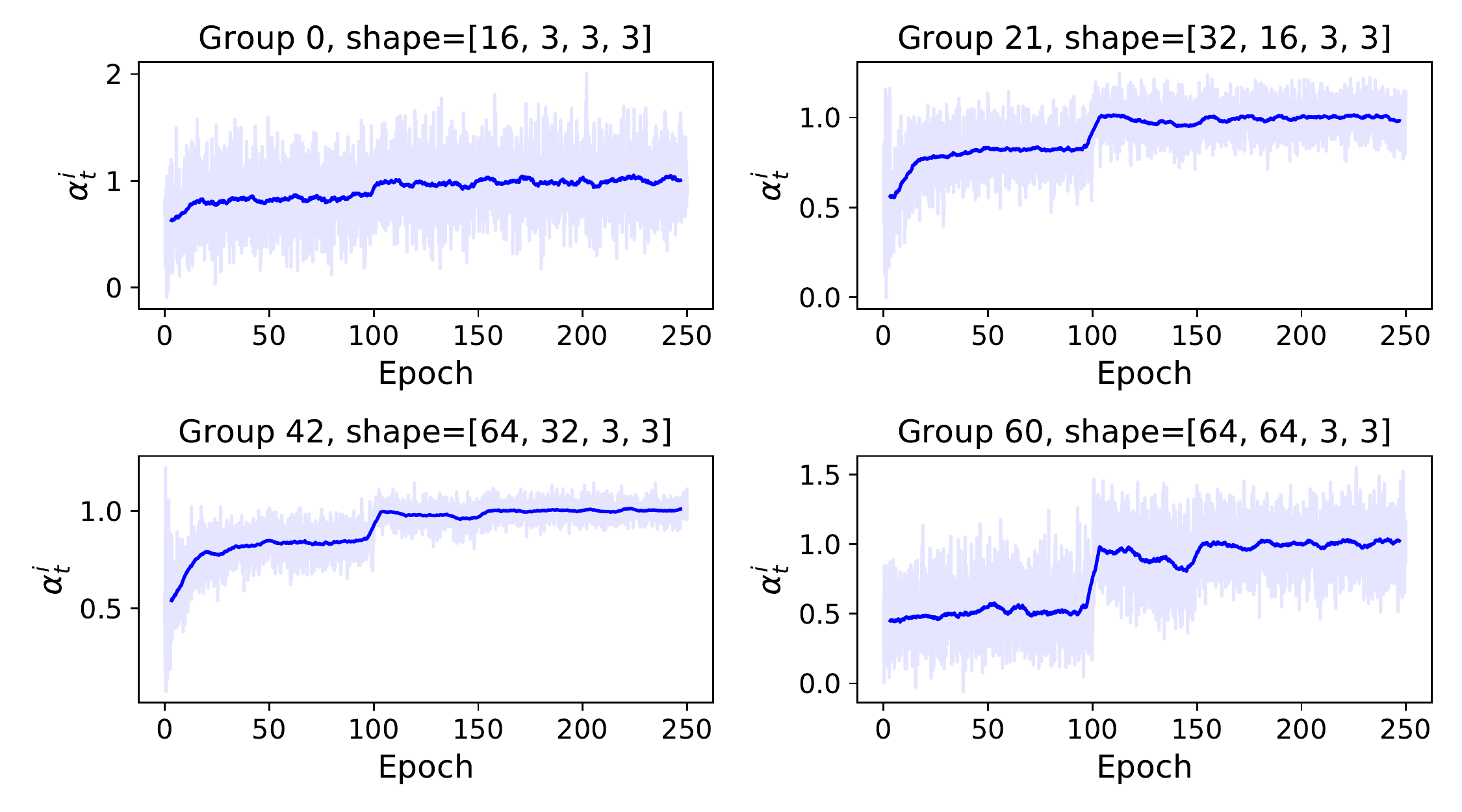}  
\vskip -0.1in
\caption{The average gradient scaling $\alpha^i_t$ increases throughout training. This plot measures $\alpha^i_t$ for four different convolutional layers in ResNet-20. The model is trained on CIFAR-10 with SGDM+AB, $\rho=2$, with a delay of $D=32$.}
\label{fig:exp4-alpha}
\end{figure}

In Figure~\ref{fig:exp4-alpha}, we measure the gradient scaling $\alpha^i_t$ applied by Adaptive Braking during ResNet-20 + CIFAR-10 training. At the start of training, successive gradients are well-aligned, so as expected $\alpha^i_t$ is less than one and AB scales down the gradients. In the later stages of training, successive gradients are not well aligned and $\alpha^i_t$ returns closer to 1, which would be equivalent to vanilla SGDM.

We also find that the gradient scaling rarely becomes negative, despite the fact that we are using a braking coefficient of $\rho=2$. This confirms that even though the potential range of the gradient scaling is $1 - \rho \leq \alpha^i_t \leq 1 + \rho$, the typical values seen during CNN training rarely reach such extreme values. This is probably due to the high dimensionality of the parameter groups, which leads the average cosine similarity to be closer to zero. 

The norm of the gradient is also usually smaller than the norm of the velocity ($\tau^i_t < 1$), so even if $\alpha^i_t$ does become negative at an individual step, it will likely only reduce the velocity norm, not completely reverse the direction of optimization (gradient ascent). In extreme cases such as a loss plane with constant gradient, we would see oscillations if $\rho > 1$. But again, we do not see this behavior in practice when training CNNs.

\section{Velocity Norm and Gradient Velocity Ratio} \label{sec:vel-norm}

\begin{figure}[t]
\centering
\includegraphics[width=\linewidth]{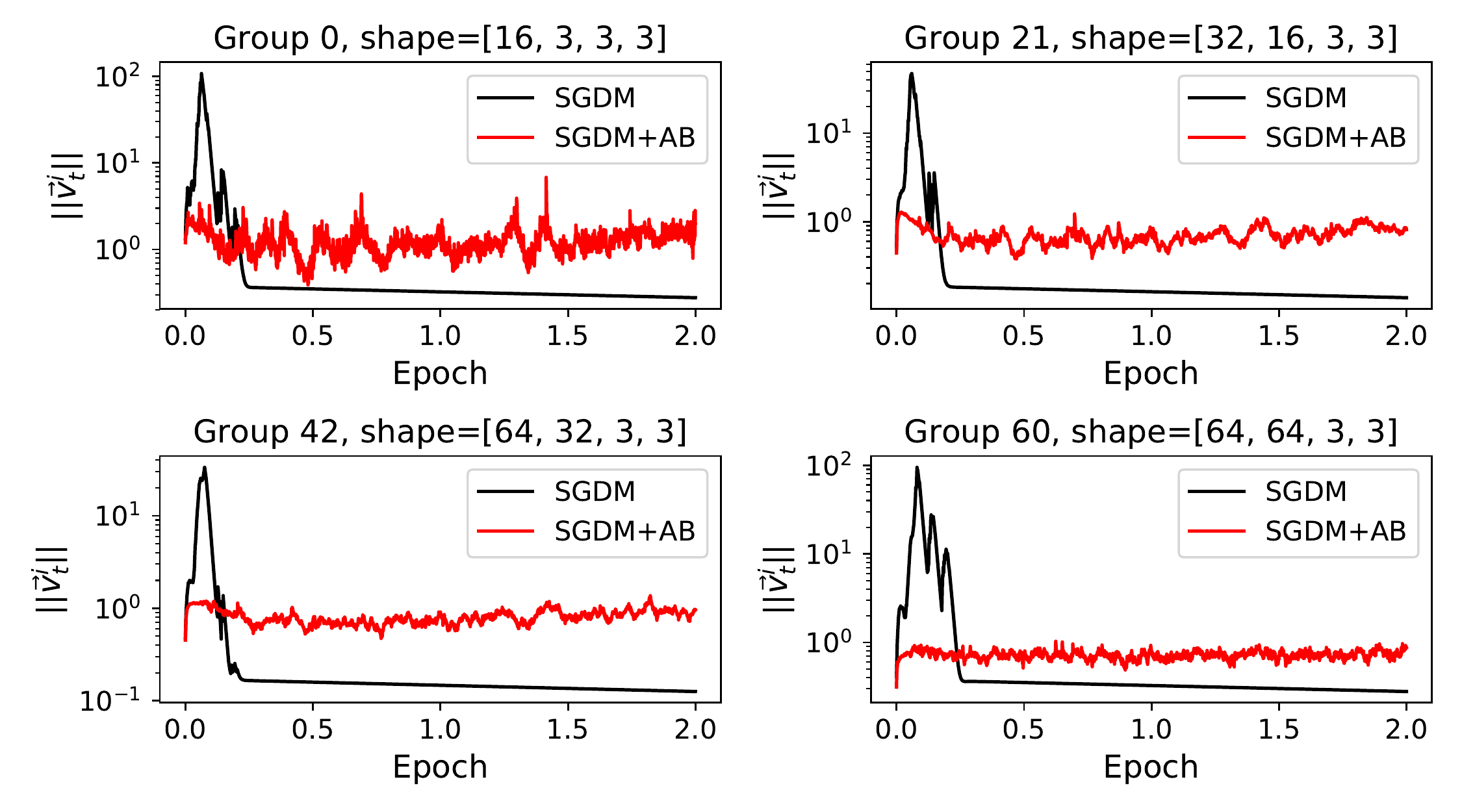}  
\vskip -0.1in
\caption{At the very start of training, SGDM+AB scales down similarly-aligned gradients, and prevents a blowup of the velocity norm that occurs with vanilla SGDM. Each plot measures $\|\vec{v}^i\|$ for a different group of convolutional layer weights $\vec{w}^i$ in ResNet-20. The y-axis is log scaled. The model is trained on CIFAR-10 with a delay of $D=32$.}
\label{fig:exp4-vel-fine}
\end{figure}

AB tends to reduce or remove the growth in the velocity that happens if successive gradients are well aligned. This is a pervasive problem in delayed gradient training, especially at the very start of training where the first $\approx D$ gradient estimates are computed on the same initial weights. In Figure~\ref{fig:exp4-vel-fine}, we plot a fine-grained view of $\|\vec{v}^i_t\|$ at the very start of ResNet-20 + CIFAR-10 training. We train with either SGDM (black) or SGDM+AB (red) and use a constant delay of $D=32$.  Without AB, the velocity norm across many parameter groups explodes within the first few hundred steps. This means that the gradients are well aligned and sum constructively, leading to large $\|\vec{v}^i\|$. When AB is used, this initial blowup is greatly reduced.

\begin{figure}[t]
\centering
\includegraphics[width=\linewidth]{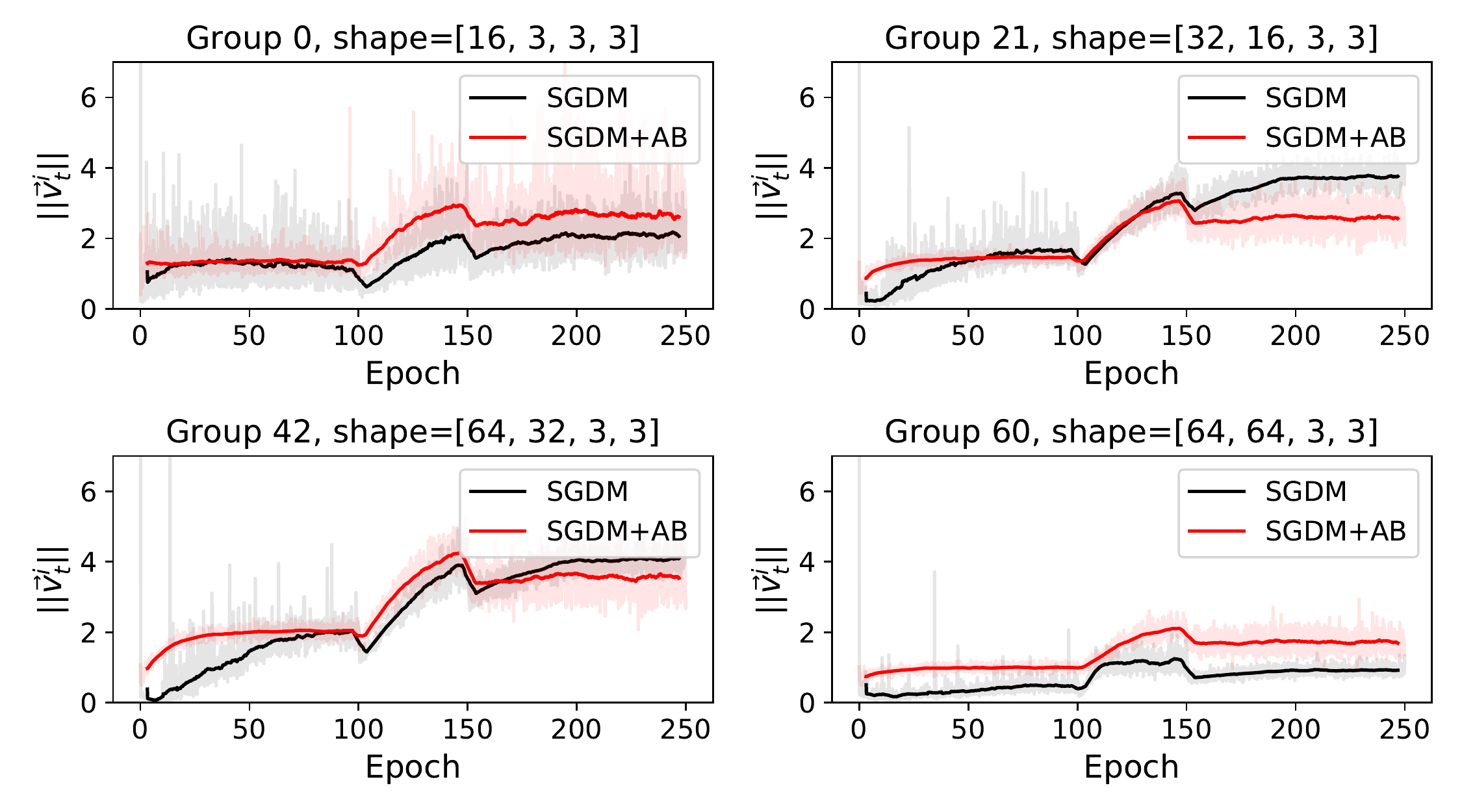}  
\vskip -0.1in
\caption{The velocity norm $\|\vec{v}^i_t\|$ measured across a full training run, for four different convolutional layers in ResNet-20. During early training, the velocity norm is larger when using SGDM+AB than when using vanilla SGDM. The model is trained on CIFAR-10 with a delay of $D=32$.}
\label{fig:exp4-vel}
\end{figure}

As training continues, we notice that the average magnitude of $\|\vec{v}^i_t\|$ is not very different between SGDM and SGDM+AB, and can often be higher when using SGDM+AB (See Figure~\ref{fig:exp4-vel}). So even though Adaptive Braking is scaling down the gradient, and limiting the growth of the velocity, AB does not significantly decrease the average velocity norm $\|\vec{v}^i\|$. Instead, AB actually stabilizes the velocity and leads the optimizer to take larger steps in early training than it would with vanilla SGDM. 

This suggests that replacing Adaptive Braking with a smaller learning rate would not produce the same benefits. Slowing down the growth of the velocity vector is not the same as reducing the magnitude of the weight updates. This idea is explored further in Appendix \ref{sec:unwrap}.

To better measure the effect of AB across different layers and across the training schedule, we introduce Gradient Velocity Ratio (GVR), which measures the ratio of the gradient norm over the velocity norm for a group $i$ of parameters:

\begin{align}
    \tau^i_t = \frac{\|\vec{g}^i_t\|}{\|\vec{v}^i_t\|} \label{eq:turn}
\end{align}

\begin{figure}[t]
\centering
\includegraphics[width=\linewidth]{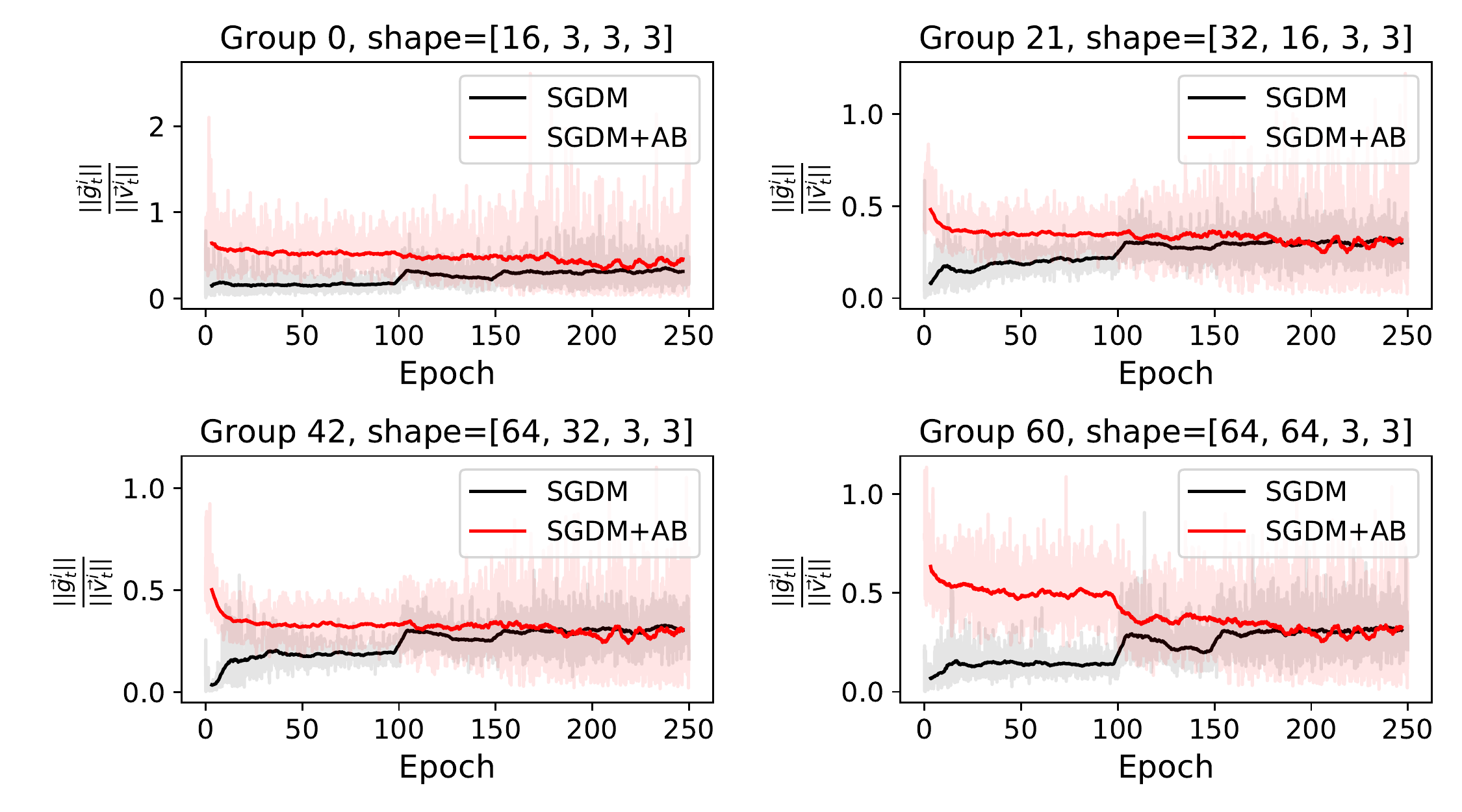}  
\vskip -0.1in
\caption{The GVR $\tau^i_t = \|\vec{g}^i_t\| / \|\vec{v}^i_t\|$ is higher when training with SGDM+AB than with vanilla SGDM. Each plot measures $\|\vec{v}^i\|$ for a different group of convolutional layer weights $\vec{w}^i$.  The model is trained on CIFAR-10 with a delay of $D=32$.}
\label{fig:exp4-turnability}
\end{figure}

We believe GVR is a good measure of a momentum-based optimizer's ability to change its trajectory. We measure the GVR $\tau^i_t$ during training in Figure~\ref{fig:exp4-turnability}, and find that using AB greatly increases GVR throughout training. This supports our theory that Adaptive Braking makes it easier to change the direction of the optimization trajectory during ASGD training.

\section{Weight Update Direction}
\begin{figure}
    \centering
    \includegraphics[width=\linewidth]{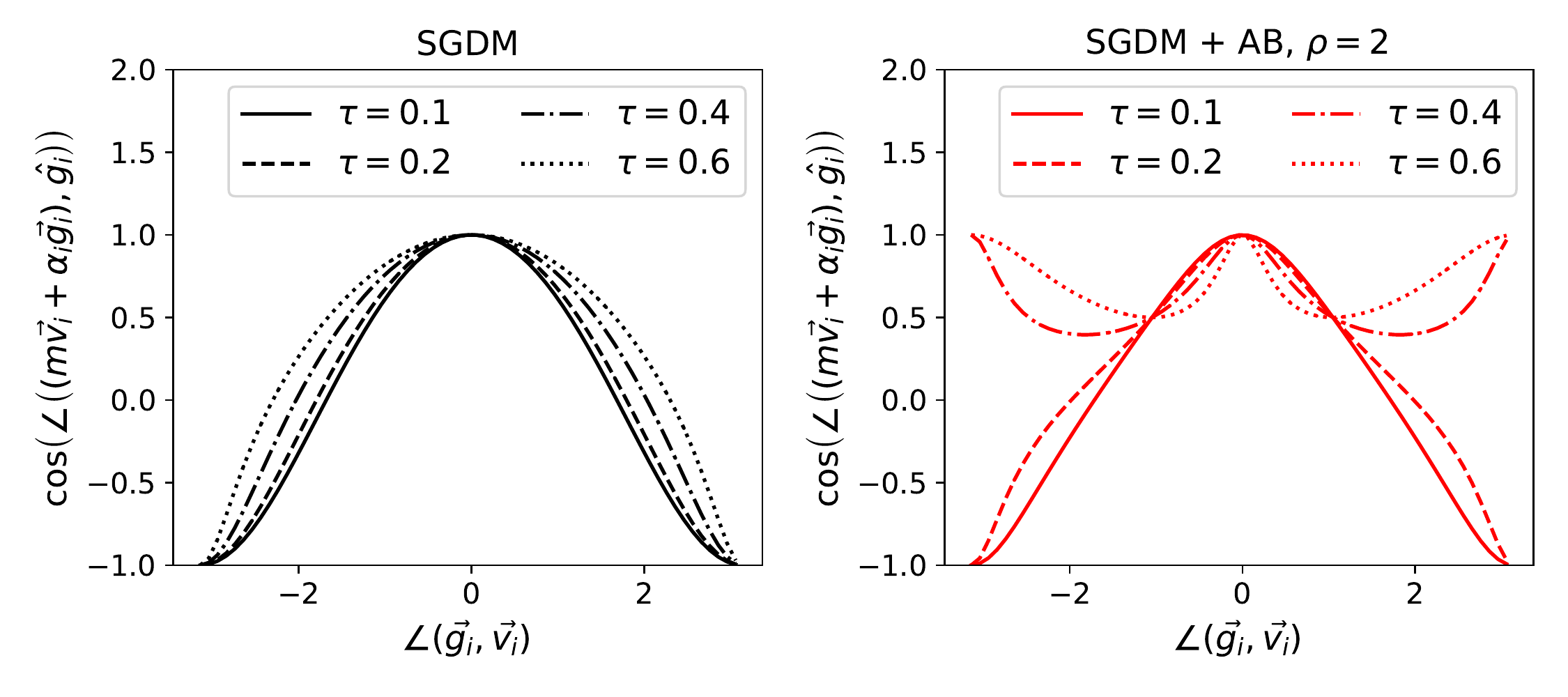}
    \vskip -0.1in
    \caption{Alignment of weight update with gradient, as a function of alignment of gradient and velocity. When the direction of $\vec{g}^i$ disagrees with the direction of $\vec{v}^i$, the SGDM + AB weight update is closer to $\vec{g}^i$.}
    \label{fig:step_direction}
\end{figure}

The instantaneous effect of AB on the SGDM weight update is to make the weight update more aligned with the gradient when the gradient and velocity directions disagree. This effect is illustrated in Figure~\ref{fig:step_direction}, where we compare the alignment of the gradient with the weight update for both SGDM and SGDM+AB, $\rho=2$. We plot measurements for a range of gradient-velocity ratio (GVR) values $\tau = [0.1, 0.2, 0.5, 0.8]$ which we find is typical in CNN training (See Appendix \ref{sec:vel-norm} and Figure~\ref{fig:exp4-turnability}). 

Note that a similar effect can be achieved for vanilla SGDM if we significantly reduce the momentum $m$: for instance if $m=0$ then the weight update is always perfectly aligned with the gradient. When training with delayed gradients and no mitigation, setting $m=0$ is a valid choice and can even be optimal when there is no gradient noise. However we find that in both the convex quadratic and neural network ASGD setting that we can achieve faster convergence if we use nonzero momentum combined with delay mitigation. This is why AB's ability to reorient the weight update in the presence of momentum is valuable.

\section{AB Ablation Study} \label{sec:unwrap}
Since AB scales the gradient during both the velocity update step and the weight update step, we can ask whether the delay tolerance of AB comes from just the instantaneous correction to the weight update, or from the long-term effect on the velocity. The two effects are made clear by looking at an unwrapped version of the SGDM+AB update equations:

\begin{align}
\label{eq:ab-vel-update}
\vec{v}_{t+1}^i &= m \vec{v}_{t}^i + \alpha_{t}^i \vec{g}_{t}^i \\
\label{eq:ab-weight-update}
\vec{w}_{t+1}^i &= \vec{w}_{t}^i - \eta \left( m \vec{v}_{t}^i + \alpha_{t}^i \vec{g}_{t}^i \right)
\end{align}

For SGDM, the scaling factor $\alpha_t^i$ is always fixed to 1. For SGDM+AB, $\alpha_t^i$ is normally computed per-step as described by \eqref{eq:ab_alpha} and applied in both equations. Alternatively, we can apply the scaling in only one equation or the other. 

If we choose to apply gradient scaling only on the velocity update, we call this algorithm AB-vel-only, with update equations:

\begin{align}
\label{eq:ab-vel-only-1}
\vec{v}_{t+1}^i &= m \vec{v}_{t}^i + \alpha_{t}^i \vec{g}_{t}^i \\
\label{eq:ab-vel-only-2}
\vec{w}_{t+1}^i &= \vec{w}_{t}^i - \eta \left( m \vec{v}_{t}^i + \vec{g}_{t}^i \right)
\end{align}

If we choose to apply gradient scaling only on the weight update, we call this algorithm AB-weight-only, with update equations:

\begin{align}
\label{eq:ab-step-only-1}
\vec{v}_{t+1}^i &= m \vec{v}_{t}^i + \vec{g}_{t}^i \\
\label{eq:ab-step-only-2}
\vec{w}_{t+1}^i &= \vec{w}_{t}^i - \eta \left( m \vec{v}_{t}^i +  \alpha_{t}^i \vec{g}_{t}^i \right)
\end{align}

\begin{figure}[t]
\centering
\includegraphics[width=\linewidth]{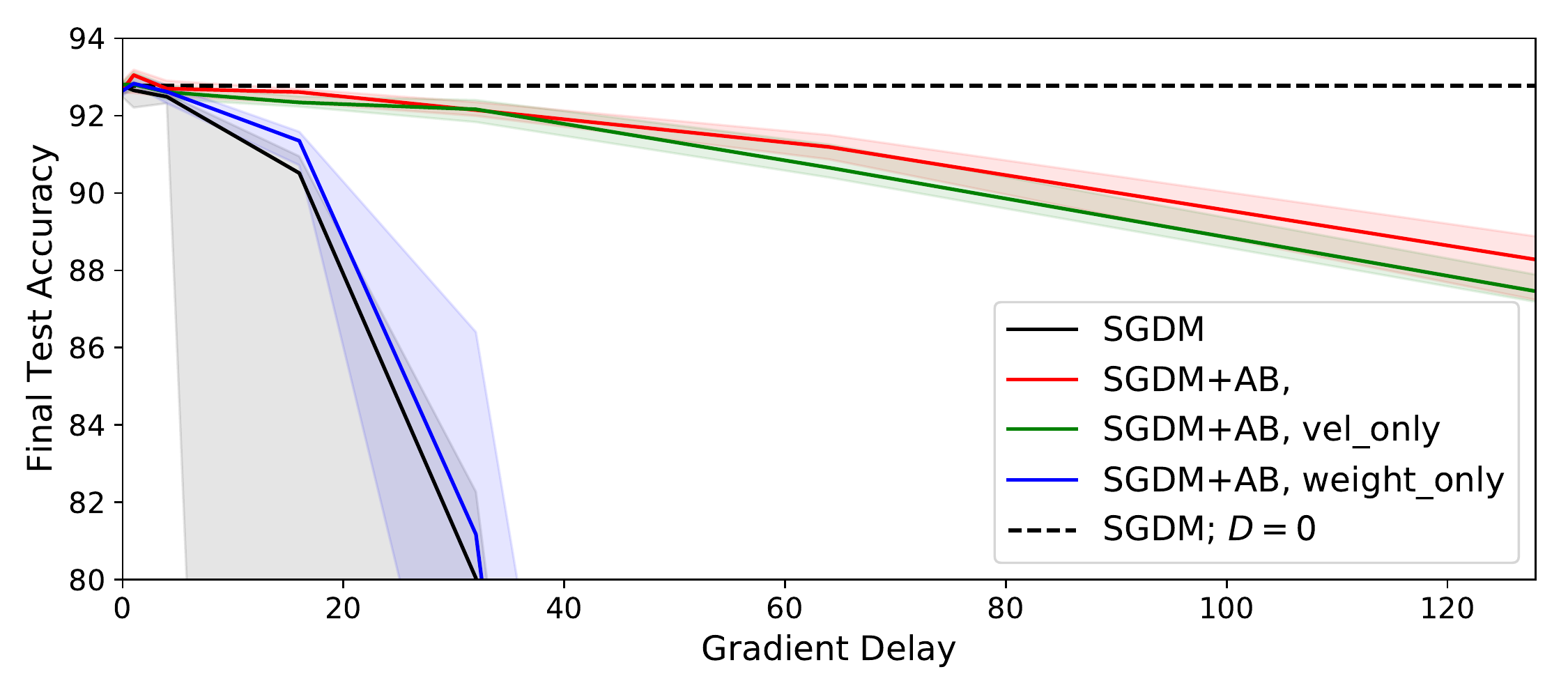}
\vskip -0.1in
\caption{Scaling the gradient during the velocity update is more important for delay tolerance than scaling the gradient during the weight update.}
\label{fig:exp5-vel-step}
\end{figure}

In Figure~\ref{fig:exp5-vel-step}, we measure the delay tolerance of SGDM with either AB-vel-only or AB-weight-only. We find that most of the delay tolerance of AB comes from scaling the gradient before updating the velocity. This supports the theory that balancing the velocity norm and dampening oscillations is crucial to mitigating delays.

\section{Extended Noisy Quadratic Model Analysis}
\begin{figure*}[t!] 
    \vskip -0.05in
    \centering
    \includegraphics[width=0.8\linewidth]{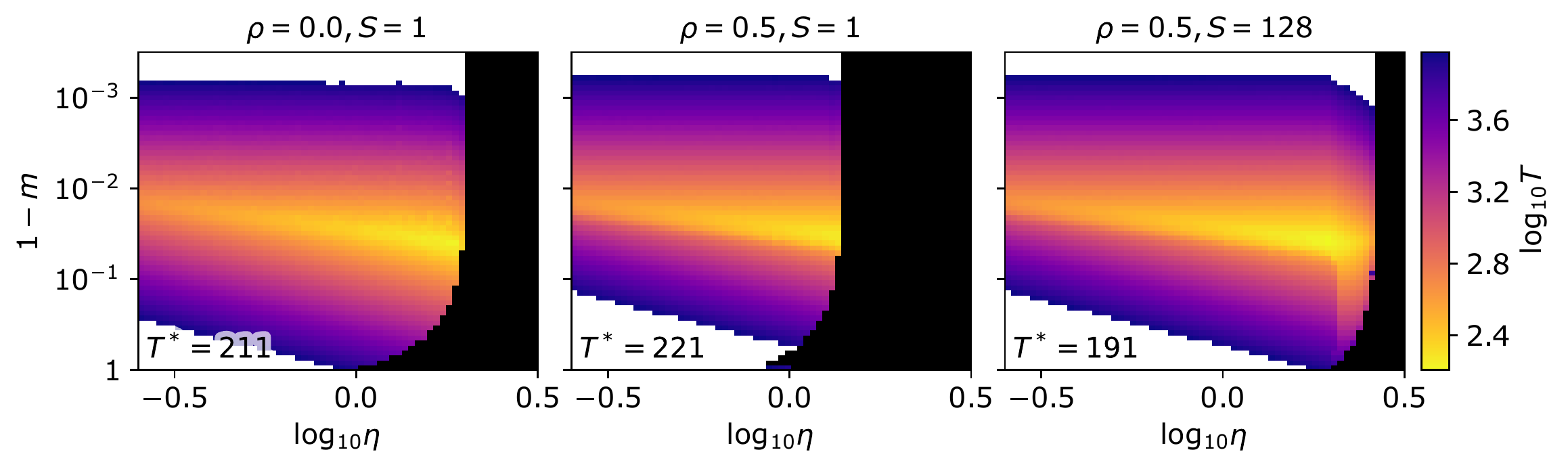}
    \vskip -0.175in
    \caption{
    This figure shows the effect of micro-stepping on the convergence region in the first row of Figure~\ref{fig:heatmap}. Only high learning rates are shown and 10000 steps are performed. With micro-stepping AB increases the stability region compared to plain SGDM and unlocks faster trajectories.
    }
    \vskip -0.1in
    \label{fig:heatmap_microstep}
\end{figure*}

\subsection{Problem Setup} \label{sec:nqm_setup}
We assume that the convex quadratic is centered and aligned with the axis. This can be done without a loss of generality since the optimizers considered are both translation and rotation-invariant\footnote{To make AB rotation-invariant we use a single group spanning all parameters. We investigate different groupings in Appendix \ref{sec:param-grouping}.}. We write the loss as:
\begin{equation}
    \mathcal{L}(\vec{w}) = \frac{1}{2} \vec{w}^T \vec{H} \vec{w} = \frac{1}{2} \sum_{k=1}^{N} \lambda_k w_k^2
\end{equation}
where $\vec{w} = [w_1, ..., w_N]^T$ are the weights to be optimized and $\vec{H}=\textrm{diag}(\lambda_1, ..., \lambda_N)$ is the Hessian of the loss.

Following \citeauthor{zhang2019algorithmic}~\citeyearpar{zhang2019algorithmic} we assume additive gradient noise with covariance equal to the Hessian of the loss. We also adopt their Hessian eigenvalue spectrum which is of the form $\{\frac{1}{j}\}_{j=1}^{N}$ with $N=10^4$ which they show can closely match certain neural networks. We write the gradient at timestep $t$ as:
\begin{equation}
    \vec{g}_t = \vec{H} \vec{w}_{t-D} + \sigma \mathcal{N}(\vec{0}, \vec{H})
\end{equation}
where $\vec{w}_{t-D}$ are the weights with delay $D$, $\mathcal{N}(\vec{0}, \vec{H})$ is the multivariate normal distribution noise with mean $\vec{0}$ and covariance matrix $\vec{H}$, and $\sigma$ scales the noise.

\citeauthor{zhang2019algorithmic}~\citeyearpar{zhang2019algorithmic} use the noisy quadratic model to explore the effects of batch size (simulated by modifying the noise scale $\sigma$) with good predictive results for neural networks. Their focus is on linear optimizers which allows them to derive closed form solutions for the convergence. Since AB is non-linear and has a cross-feature dependency we explicitly carry out the optimization on the full quadratic and do not use any sort of binning of similar eigenvalues. The objective of the optimization is to bring the loss below the target loss $\varepsilon=0.01$ and the weights are initialized to $\vec{w}=\vec{1}$. We measure the quality of trajectories with the number of steps, $T$, required to reach the target loss.

\subsection{Energy Measure} \label{sec:nqm_energy}
In Section~\ref{sec:nqm} we explore the effect of AB on individual components. To do this effectively we introduce an energy measure to estimate the convergence of individual components and compare it between states. Using the loss for this is problematic because it oscillates and a low loss does not necessarily indicate convergence (if the velocity is large). In more realistic settings we can not easily determine what the components are and therefore can not compute component losses, apply different learning rates to different components or early stop individual components.
We use a similar energy model as \citeauthor{hermans2018gradient}~\citeyearpar{hermans2018gradient} that accounts for both the loss (potential energy) and velocity (kinetic energy). Our energy ($E$) is normalized with the learning rate ($\eta$) making it directly comparable with the loss:
\begin{align}
    E_t &= \mathcal{L}(\vec{w}_t) + \frac{1}{2}\frac{\|\vec{w}_t-\vec{w}_{t-1}\|^2}{\eta} \\
        &= \mathcal{L}(\vec{w}_t) + \frac{1}{2}\eta \|\vec{v}_t\|^2
\end{align}

Note that the energy upper bounds the loss so an energy of zero would mean that a component has fully converged.
For an oscillating trajectory the energy is roughly equal to the loss at the extreme points where the velocity is approximately zero.
Overall the energy can be viewed as roughly estimating the envelope of the loss for an oscillating component.
This makes it easier to estimate convergence from a single state and compare the convergence of different states than using the loss directly.

\begin{algorithm}[h!]
\caption{AB with micro-stepping}
\label{alg:AB_microstepping}
\begin{algorithmic}
    \setlength{\itemindent}{-0.5em} 
    \STATE For t = 1...T do:
    \STATE \quad Compute gradient $\vec{g}$
    \STATE \quad $\vec{v} \gets m \vec{v}$
    \STATE \quad For i = 1...S do:
    \STATE \quad \quad $\alpha \gets 1 - \rho \frac{\langle \vec{g}, \vec{v} \rangle}{\max(\|\vec{g}\| \|\vec{v}\|, \epsilon)}$
    \STATE \quad \quad $\vec{v} \gets \vec{v} + \frac{\alpha}{S}\vec{g}$
    \STATE \quad $\vec{w} \gets \vec{w}-\eta \vec{v}$
\end{algorithmic}
\end{algorithm}

\begin{figure*}[t!] 
    \vskip -0.05in
    \centering
    \includegraphics[width=1.0\linewidth]{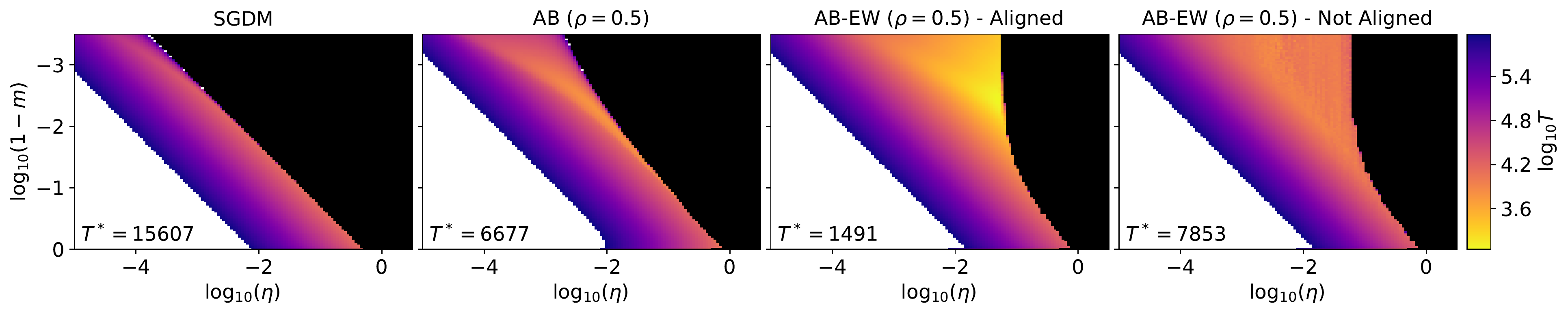}
    \vskip -0.175in
    \caption{
    Convergence heatmaps for the optimization of a low dimensional convex quadratic (CQ) with a delay of one.
    Plain SGDM and standard AB are rotation invariant but applying AB element-wise is not.
    Element-wise AB performs very well if the axes of the CQ are aligned, such that the AB is applied to each component individually (third panel).
    The last panel shows element-wise AB for a CQ with a random alignment.
    In this case applying AB element-wise does not work as well as the global form, although the region of stability is larger.
    }
    \vskip -0.1in
    \label{fig:heatmap_abew}
\end{figure*}

\subsection{Micro-stepping} \label{sec:nqm_microstepping}
Figure~\ref{fig:heatmap} shows that in the no-delay and no-noise case AB can slightly reduce the region of stability. This leads to sightly worse optimal trajectories. In Section~\ref{sec:nqm} we state that this happens because AB can magnify certain high frequency oscillations. With noise and delays this does not seem to be an issue, potentially because the baseline SGDM trajectories don't converge to the target loss for hyperparameter settings where high frequency oscillations could occur.

As an example of AB magnifying oscillations, consider the case where AB is applied on a single component with curvature $\lambda$, learning rate $\frac{1}{\lambda} < \eta < \frac{2}{\lambda}$ and very small momentum value $m \approx 0$. This will result in a trajectory that overshoots the minimum at every step and $\vec{g}$ and $\vec{v}$ will always be oppositely aligned. This causes AB to apply a constant $\alpha=1+\rho$, effectively increasing the learning rate, potentially causing instability. 

The issue arises from AB over-correcting the velocity when the gradient $\vec{g}_t$ and velocity $\vec{v}_t$ are oppositely aligned. This happens because AB scales the gradient based on the alignment of $\vec{v}_t$ and $\vec{g}_t$ without considering the resulting alignment of $\vec{g}_t$ and $\vec{v}_{t+1}$. In cases where $\|\vec{v}_t\|$ is small and $\vec{g}_t$ and $\vec{v}_{t+1}$ are oppositely aligned this can lead to larger $\|\vec{v}_{t+1}\|$. Various forms of clamping can help here, for example enforcing $\alpha \le 1$ but we have found that this can reduce the effectiveness of AB.

Another way is to change the velocity update to consider more than just the initial alignment of $\vec{g}_t$ and $\vec{v}_t$. We can divide the velocity update into $S$ ``micro-steps", calculating a different $\alpha$ for each one as shown in Algorithm~\ref{alg:AB_microstepping}. For large values of $S$ micro-stepping might have significant overhead but could help AB in the large batch size or low noise settings. Figure~\ref{fig:heatmap_microstep} shows the effects of micro-stepping on the speed of convergence. It shows that with micro-stepping AB can tolerate higher learning rates than plain SGDM and slightly decreases the minimum steps needed to reach the target loss.

\subsection{Parameter Grouping} \label{sec:nqm_grouping}
Adaptive Braking operates by computing an alignment score between the gradient and velocity for a group of parameters and then scaling the gradient based on the alignment. The performance of AB depends on the choice of groups. For neural networks we find that filter-wise grouping works well, see Appendix~\ref{sec:param-grouping}. In this section we explore the effect of grouping for convex quadratics, in particular we compare the global form (with a single group) to the element-wise form.

To decease compute requirements we use low dimensional models in this section. We use 32 components with a log-uniform eigenvalue spectrum from $10^{-4}$ to $1$ and a target loss of $\epsilon=10^{-5}$. Figure~\ref{fig:heatmap_abew} shows the steps required to reach the target loss for different AB forms for a delay of $1$ and no noise. We can see that the global form of AB outperforms the baseline. The element-wise form works really well if the quadratic aligns with the axes. In this case it is really performing component-wise AB. This can speed up the convergence of all components that are sufficiently underdamped. For overdamped components this slows their convergence (by effectively lowering the learning rate). However, since all components are stabilized, higher learning rates can be used which at least partially compensates for this effect. Ideally we could apply AB selectively to the components that need to be dampened without affecting the other ones. Unfortunately we generally don't know what the components are and element-wise AB does not necessarily outperform the global form of AB for a random alignment (see Figure~\ref{fig:heatmap_abew}).

Overall there seems to be a trade-off in the group size. Each additional component in a group lowers the correlation of the scaling to the other components, weakening the dampening effect. Using a larger number of groups, with fewer components each, may give stronger correlations increasing the dampening effect. Ideally the most unstable components should fall in separate groups so they can be dampened effectively. It may also be important for components to be contained within a single group. If this is not the case, different coordinates of the gradient for a given component may be scaled differently. This effectively rotates the gradient, potentially causing it to interfere with the convergence of other components. This might be why element-wise AB generally doesn't perform as well as using larger groups (when the loss is not aligned as is usually the case). The filter-wise grouping we use for neural networks (see Appendix~\ref{sec:param-grouping}) could strike a good balance between the number of groups and splitting components between groups.

\end{document}